\documentclass[10pt]{article}
\usepackage[accepted]{tmlr}

\usepackage{microtype}
\usepackage{graphicx}
\usepackage{subcaption}
\usepackage{booktabs} % for professional tables

\usepackage[backref=page]{hyperref}

\usepackage{times}

\usepackage[utf8]{inputenc} % allow utf-8 input
\usepackage[T1]{fontenc}    % use 8-bit T1 fonts
\usepackage{url}            % simple URL typesetting
\usepackage{booktabs}       % professional-quality tables
\usepackage{amsfonts}       % blackboard math symbols
\usepackage{amsmath}
\usepackage{amssymb}
\usepackage{mathtools}
\usepackage{amsthm}

\usepackage[capitalize,noabbrev]{cleveref}
\usepackage{mathtools}
\usepackage{thmtools}       % fixes autoref with thm
\usepackage{nicefrac}       % compact symbols for 1/2, etc.
\usepackage{xcolor}         % colors
\usepackage{algorithm, algpseudocode}
\usepackage[english]{babel}
\usepackage{acronym}
\usepackage{graphicx}
\usepackage[backref=page]{hyperref}       % hyperlinks
\hypersetup{
    colorlinks=false,
    citebordercolor=cyan,
    linkcolor=blue,
    filecolor=blue,      
    citecolor=blue,
}
\usepackage[font=small]{caption}
\usepackage[font=small]{subcaption}
\usepackage{letltxmacro}
\acrodef{RL}{reinforcement learning}
\acrodef{SL}{supervised learning}
\acrodef{TSCL}{Teacher-Student Curriculum Learning}

\acrodef{MDP}{Markov Decision Process}

\acrodef{AI}{Artificial Intelligence}
\acrodef{ML}{Machine Learning}
\acrodef{vPoP}{Value of a Player to other Player}
\acrodef{A-SIPD}{\textit{Adversarial Sparse Iterated Prisoner's Dilemma}}
\acrodef{PER}{Prioritized Experience Replay}
\newtheorem{theorem}{Theorem}[section]

\theoremstyle{definition}
\newtheorem{definition}{Definition}[section]
\newtheorem{example}[theorem]{Example}

\theoremstyle{remark}

\addto\extrasenglish{

}

\title{Rethinking Teacher-Student Curriculum Learning through the Cooperative Mechanics of Experience}

\author{\name Manfred Diaz \email diazcabm@mila.quebec \\
      \addr Mila, University of Montreal \\
      \AND
      \name Liam Paull \email paulll@mila.quebec \\
      \addr Mila, University of Montreal \\
      Canada CIFAR AI Chair
      \AND
      \name Andrea Tacchetti \email attachet@google.com \\
      \addr Google DeepMind
      }

\def\month{09}  % Insert correct month for camera-ready version
\def\year{2024} % Insert correct year for camera-ready version
\def\openreview{\url{https://openreview.net/forum?id=qWh82br6KT}} % Insert correct link to OpenReview for camera-ready version

\begin{document}
% \normalsize

\maketitle

\begin{abstract}
Teacher-Student Curriculum Learning (TSCL) is a curriculum learning framework that draws inspiration from human cultural transmission and learning. It involves a teacher algorithm shaping the learning process of a learner algorithm by exposing it to controlled experiences. Despite its success, understanding the conditions under which TSCL is effective remains challenging. In this paper, we propose a data-centric perspective to analyze the underlying mechanics of the teacher-student interactions in TSCL. We leverage cooperative game theory to describe how the composition of the set of experiences presented by the teacher to the learner, as well as their order, influences the performance of the curriculum that is found by TSCL approaches. To do so, we demonstrate that for every TSCL problem, an equivalent cooperative game exists, and several key components of the TSCL framework can be reinterpreted using game-theoretic principles. Through experiments covering supervised learning, reinforcement learning, and classical games, we estimate the cooperative values of experiences and use value-proportional curriculum mechanisms to construct curricula, even in cases where TSCL struggles. The framework and experimental setup we present in this work represents a novel foundation for a deeper exploration of TSCL, shedding light on its underlying mechanisms and providing insights into its broader applicability in machine learning.
 \end{abstract}

\section{Introduction}

Controlling the sequence of tasks that a learning algorithm is exposed to through curriculum has been shown to potentially enhance learning efficiency~\citep{elman1993curriculum, krueger_dayan2009curriculum, bengio2009curriculum}. One widely used curriculum framework, known as \ac{TSCL}~\citep{graves2017curriculumbandit, matiisen2020curriculumbandit}, specifically gives a \textit{teacher algorithm} the ability to control this sequence. 
While it is commonly understood that presenting tasks with increasing difficulty can improve learning, the underlying dynamics and structure of teacher-student interaction in this context are still relatively unexplored.
Very few works have attempted to understand \textit{when}, and \textit{how}~\ac{TSCL} works~\citep{ Lee_Goldt_Saxe_2021, Wu2020-hq} while most have focused on providing algorithmic improvements to the problem~\citep{portelas2019curriculumbandit, turchetta2020curriculumbandirsaferl, liu2020curriculumbanditadam, feng2021curriculumbandit}. 
%Liam's version of the below paragraph:
In this paper, we propose a novel \emph{data-centric} perspective~\citep{ng2021data} to understand and analyze \ac{TSCL} algorithms. 

We begin by formalizing a general notion of \textit{units of experience} to describe the teacher algorithm's control objects (consumed by the learner). Subsequently, our approach draws inspiration from work on feature attribution~\citep{romapatel2021deepmind}, data valuation~\citep{ghorbani2019datashapley, yanprocaccia2021shapleylovecore} and explainability~\citep{lundberg2017shap},
and leverages tools from cooperative game theory~\citep{vonneumann1944theory, shapley1952value} 
to analyze \textit{how} the compositions of these units impact teacher-student interactions. 
% In this paper, we propose a novel \emph{data-centric} perspective~\citep{ng2021data} to understand \ac{TSCL} that broadens this framework's scope to describe a family of curriculum learning algorithms that learns to prioritize, not just a set of tasks, but a set of \emph{units of experience} (e.g., a set of \textit{instances}, or \textit{environments}, or \textit{tasks}) (\autoref{sec:experience-to-control}), 
% and draws inspiration from work on \textit{feature attribution}~\citep{romapatel2021deepmind}, \textit{data valuation}~\citep{ghorbani2019datashapley, yanprocaccia2021shapleylovecore} and \textit{explainability}~\citep{lundberg2017shap},
% and leverages tools from cooperative game theory~\citep{vonneumann1944theory, shapley1952value} 
% to explain \textit{how} the composition of this units set conditions teacher-student interactions. 
We show that, for every~\ac{TSCL} problem,  
there exists an equivalent cooperative game where \textit{units of experience} are players and teacher-student interactions approximate a sequential coalition formation process (\autoref{sec:cooperative-game}). As a result, the learning progression objective~\citep{Schmidhuber_1991, oudeyer2007intrinsic, graves2017curriculumbandit} and the teacher bandit policy~\citep{gittins1979bandits, matiisen2020curriculumbandit}, two essential components of \ac{TSCL},
have alternative interpretations as an approximation of player (unit) marginal contribution~\citep{ weber1988probshapley} and a fair allocation mechanism, respectively (\autoref{sec:marginal-contribution} \& \ref{sec:fair-allocation}).
% Liam: I think we need to introduce the traditional CL game-theoretic arguments first here so that the reader can follow
% Manfred: I am not aware of any in this line of work
% Traditional game-theoretic formulations of CL model the problem as BLAHBLAH game or something or other (FILL I?N). 
Furthermore, because the \textit{order matters} in the case of curriculum learning~\citep{krueger_dayan2009curriculum, bengio2009curriculum}, traditional cooperative game-theoretic arguments produce unintuitive results~\citep{Nowak1994shapleygen}. Thus, we leverage \textit{generalized cooperative games} and their solution concepts~\citep{Nowak1994shapleygen, Sanchez1997shapleygen} to overcome these limitations and formally extend these data-centric game-theoretic formulations to the curriculum learning setting.

To demonstrate the predictive power and range of problems where this game-theoretic and data-centric interpretation of~\ac{TSCL} applies, we build an experimental setting that evaluates the prospect of cooperation among \textit{units 
of experience} in problems spanning \ac{SL}, \ac{RL}, and classical games (\autoref{sec:expensive-prior}).
These experiments simulate ordered and unordered coalition formation processes and approximate the cooperative games we developed to describe~\ac{TSCL}. For every problem, we estimate units \emph{a priori} value (e.g., \citeauthor{shapley1952value} or \citeauthor{Nowak1994shapleygen} values) and demonstrate that these \emph{a priori} values, although expensive to compute~\citep{deng_papadimitriou1994complexitycoop}, are useful proxies to find curricula. To this end, we design unordered and ordered value-proportional curriculum mechanisms inspired by value-proportional allocations~\citep{bachrach2020teamformation}. In most settings, the unordered mechanism fails to find a reasonable curriculum, demonstrating the unsuitability of traditional game-theoretic tools for the~\ac{TSCL} problem. However, the ordered mechanism consistently finds an optimal or near-optimal ordering (i.e., a curriculum) even when \ac{TSCL} fails (\autoref{sec:values-curriculum}).  
To understand what impacts the ability of~\ac{TSCL} in those settings, we leverage another cooperative game-theoretic tool, namely, \emph{measures of interactions}~\citep{Grabisch_Roubens_1999, Procaccia_Shah_Tucker_2014}, and in particular the \ac{vPoP}~\citep{hausken_mohr_2001shapleyothers}, to quantify positive, neutral, or negative pairwise interactions among units. We show that in settings with considerable unit interference, as characterized by their negative pairwise interactions, \ac{TSCL} cannot produce useful curricula.

%The framework and experimental setup we present here are a starting point for more thorough explorations of \ac{TSCL} and curriculum learning, their underlying mechanisms and broader applicability in machine learning.

\section{Preliminaries}
\label{sec:background}

\subsection{Cooperative Game Theory}
\label{sec:background-games}
% We investigate two interpretations of teacher-student interactions as a \textit{sequential coalition formation process}~\citep{Shapley1953valueforward} leading to two models of cooperative games.
\textbf{Cooperative Games.} Cooperative games model problems where players interact to maximize collective gain~\citep{roth1988shapley}. 
In a (traditional) cooperative game in characteristic function form among a set of players $\textbf{U}$, denoted by $\mathcal{G}=\left<\textbf{U}, v\right>$, the characteristic function $v: 2^\textbf{U} \to \mathbb{R}$ associates to each coalition $\textbf{C} \in 2^\textbf{U}$, belonging to the powerset $2^{\textbf{U}}$, a real number that represents the benefits produced by the players in $\textbf{C}$ acting jointly.  
In a cooperative game, a solution concept represents a mechanism that produces allocation vectors $\phi \in \mathbb{R}^{|\textbf{U}|}$~\citep{shubik1981solutions}. 
Particularly, \textit{Shapley's value}~\citep{shapley1952value} allocates to each player $\textbf{u} \in \textbf{U}$ its average marginal contribution $v(\textbf{C} + \textbf{u}) - v(\textbf{C} )$ to coalitions $\textbf{C} \subseteq \textbf{U}$, where $\textbf{u} \in \textbf{U} - \textbf{C}$
\begin{equation}
    \phi(\textbf{u}) = \sum_{\textbf{C}: \textbf{u} \not{\in} \textbf{C}} \frac{|\textbf{C}|!(|\textbf{U}| - |\textbf{C}| - 1)!}{|\textbf{U}|!} \left[ v(\textbf{C} + \textbf{u}) - v(\textbf{C}) \right] \label{eq:shapley}
\end{equation}
and uniquely satisfies the axioms of \textit{efficiency}, \textit{null-player}, \textit{symmetry}, and \textit{linearity}, which are generally considered to be properties of a fair allocation mechanism~\citep{vandebnrink1998shapleyaxioms}. 

\textbf{Generalized Cooperative Games.} 
When the order in which players join determines coalitional worth, traditional cooperative games and their solution concepts (e.g., Shapley's value) may produce unintuitive allocations~\citep{Nowak1994shapleygen}. 
In these games, the generalized characteristic function $v: \mathcal{P}(2^\textbf{U}) \to \mathbb{R}$ assigns to every ordered coalition $\textbf{C} \in \mathcal{P}(2^\textbf{U})$ in the powerset of permutations $\mathcal{P}(2^\textbf{U})$ its worth if members join in the permutation order.
\citet{Nowak1994shapleygen} and \citet{Sanchez1997shapleygen} extended Shapley's work and proposed solution concepts for these generalized cooperative games. We focus on the former due to its intuitive formulation
\begin{equation}
    \phi_{\textsc{NR}}(\textbf{u}) = \frac{1}{|\textbf{U}|!}\sum_{\substack{\textbf{C} \in \mathcal{P}(2^\textbf{U}) \\ \textbf{C}: \textbf{u} \not\in \textbf{C}}}  \left[v(\textbf{C}:\textbf{u}) - v(\textbf{C})\right]~\label{eq:nowak_radzik}
\end{equation}
that averages, for all ordered coalitions $\textbf{C} \in \mathcal{P}(2^\textbf{U})$ where the unit $\textbf{u} \in \textbf{U}$ is appended last, its marginal contribution to the newly formed ordered coalition $\textbf{C}:\textbf{u}$.

% is said to be in \textit{generalized characteristic function form}~\citep{Nowak1994shapleygen}.

% However, for games in \textit{generalized characteristic function form}, Shapley's value produces intuitively unfair allocations ~\citep{Nowak1994shapleygen}. To that end,

\textbf{Measures of Interactions.} In a cooperative game, a measure of interaction~\citep{Grabisch_Roubens_1999, Procaccia_Shah_Tucker_2014} computes players' influences on other players' outcomes. In particular, we leverage the \emph{value of a player to another player} (vPoP)~\citep{hausken_mohr_2001shapleyothers}. For the games above, \ac{vPoP} constructs a matrix whose entries $\phi(\textbf{u}_i, \textbf{u}_j) \in \mathbb{R}$ measure the influence player $\textbf{u}_i$ exerts over player $\textbf{u}_j$. It measures how the Shapley value of a unit changes in the absence of another. More precisely, 
\begin{equation}
    \phi(\textbf{u}_i, \textbf{u}_j) = \sum_{\substack{\textbf{C} \subseteq \textbf{U} \\ \textbf{u}_i, \textbf{u}_j \in \textbf{C}}}\frac{(|\textbf{U}| - |\textbf{C}|)(|\textbf{C}| -1)!}{|\textbf{U}|!} \left[ \phi(\textbf{u}_j, \textbf{C}) - \phi(\textbf{u}_j, \textbf{C} - \textbf{u}_i) \right]\label{eq:vpop}
\end{equation}
 where $\phi(\textbf{u}_j, \textbf{C})$ is the Shapley value  of unit $\textbf{u}_j$  (\autoref{eq:shapley}) in the cooperative game restricted to players in $\textbf{C}$. This matrix marginal $\phi(\textbf{u}_i) = \sum_j \phi(\textbf{u}_i, \textbf{u}_j)$ corresponds to each player's Shapley value.
 We extend \ac{vPoP} to games in generalized characteristic function form by applying~\autoref{eq:vpop} \textit{mutatis mutandis} using ~\citet{Nowak1994shapleygen} value to provide an ordered pairwise interaction metric $\phi_{\textsc{NR}}(\textbf{u}_i, \textbf{u}_j)$.

\subsection{Bandit Algorithms}
\label{sec:background-bandits}

Multi-armed bandit algorithms provide a solution to problems of decision-making under uncertainty ~\citep{gittins1979bandits, lattimore2020bandits} where, at each interaction, a decision must be made about which arm $\textbf{u} \in \textbf{U}$ must be pulled. We are particularly interested in action-value-based algorithms that maintain empirical value estimates $q_k(\textbf{u})$ computed as
\begin{equation}
    q_k(\textbf{u}) \approx \frac{1}{N^\textbf{u}_k} \sum_{i=1}^{k-1} r(\textbf{u}_i) \mathbb{I}_{\textbf{u}_i = \textbf{u}} \label{eq:empirical_values}
\end{equation}
and that estimate the average reward received by the algorithm in the iterations $N^{\textbf{u}}_k \leq k$ where the $\textbf{u}$-arm has been pulled.
Bandit algorithms, like the ones \citet{graves2017curriculumbandit} and~\citet{matiisen2020curriculumbandit} use in their work, transform the estimated average contributions into arms interactions by deriving from estimated values a Boltzmann policy $\tau_k \in \Delta(\textbf{U})$ such that the probability of interaction is proportional to the value estimates:
\begin{equation}
    \tau_k(\textbf{u}) \propto \mathcal{B}(q_k(\textbf{u})) = \frac{e^{\frac{q_k(\textbf{u})}{T}}}{\sum_{\textbf{u}^\prime} e^{\frac{q_k(\textbf{u}^\prime)}{T}}} \label{eq:boltzmann}
\end{equation}
More sophisticated approaches (e.g., the~\textsc{Exp3}~\citep{auer2003bandits} used in our experiments) account for other factors, like recency, bias, stochasticity, or non-stationarity~\citep{lattimore2020bandits}. 

\section{Experience to Control}
\label{sec:experience-to-control}
The \ac{TSCL} framework commonly operates under the assumption that tasks presented to a learning algorithm can influence its learning dynamics. 
Modern iterative learning algorithms process tasks in discrete units. For instance, \ac{SL} and \ac{RL} algorithms operate over {instances} and {transitions}, respectively.  But also, collections of these elementary units, such as {batches} or {episodes}, {datasets} or \textit{environments}, or more generally {benchmarks} or {environment suites}, describe a hierarchy of aggregations of experience.
Henceforth, we utilize the term \emph{\textbf{unit of experience}} for referring to any collection of discrete units that a teacher algorithm can use to control the dynamics of the learner algorithm.

\begin{algorithm}[t]
    % \small
      \caption{Generalized Teacher-Student Curriculum Learning.} 
\label{alg:gentscl}
    \begin{algorithmic}[1]
        \Procedure {\textsc{GenTSCL}} {}
        \State \textbf{inputs} policy: $\pi_0$, algorithm: $\mathcal{L}$, units: $\textbf{U}$, metric: $\mathcal{J}$
        \State $\quad$ \textit{teacher}: $\tau_0 \in \Delta(\textbf{U})$,  \textit{targets}: $ \bar{\textbf{U}}$, budget: $K$
        \For{$k = 1 \ldots K$}
           \State $\textbf{u}_k \sim \tau_k(\textbf{u})$ \label{alg:teacher-policy}
           \State $\pi_{k} \sim \mathcal{L}(\pi_{k-1}, \textbf{u}_k)$ \label{alg:gentscl_control}
           \State $r_k \gets \mathcal{J}(\pi_{k}, \bar{\textbf{U}}) - \mathcal{J}(\pi_{k-1}, \bar{\textbf{U}})$ \label{alg:gentscl_learningprogression}
           \State $\tau_{k+1} \gets \textsc{UpdateRule}(\tau_k, \textbf{u}_k, r_k)$ \label{alg:tscl_bandit}
        \EndFor
        \State \textbf{output:} $\pi_{K}$
        \EndProcedure
    \end{algorithmic}

\end{algorithm}

\begin{example}
    For an analysis, we may define a \textit{unit of experience} as the set of instances of class in a \ac{SL} classification problem. For example, in the \emph{MNIST} dataset~\citep{lecun-mnisthandwrittendigit-2010}, there may be \textit{ten} \textit{units of experience}, namely, classes $\textsc{zero}, \textsc{one}, \textsc{two}, \ldots, \textsc{nine}$.
    \label{ex:mnist}
\end{example}
The \textit{units of experience} abstraction indistinctly applies to supervised or reinforcement learning problems. On either paradigm, any iterative learning algorithm 
is a controllable system whose control inputs are units of experience. 
% $\textbf{u}_k \in \textbf{U}$
\begin{example}
     There are four control inputs in mini-batch gradient descent~\citep{goodfellow2016deeplearningbook}: the mini-batch $\left\{x_1, \ldots, x_B\right\}$, the loss function $\ell$, the parameters $\theta$, and the learning rate $\eta$ such that:
        $$\theta_{k+1} = \theta_k - \eta \nabla_{\!\theta_k}\sum_{i=1}^B \ell(\theta_k, x_i)$$
\end{example}
A \ac{TSCL}-style algorithm, as presented in Alg. \ref{alg:gentscl}, solves a data-centric control problem. The \textit{learner} algorithm $\mathcal{L}(\pi_{k-1}, \textbf{u}_k)$ is a \emph{black-box} system (line \ref{alg:gentscl_control}) controlled by a \textit{teacher} algorithm  through units drawn with probability $\textbf{u}_k \sim \tau_k(\textbf{u})$. The \textit{learner} output, at each iteration $k$, is policy or model $\pi_k$ whose performance is measured by a \textit{metric function} $\mathcal{J}$ that quantifies the model's performance on a set of evaluation units $\bar{\textbf{U}}$. The teacher aims
to maximize the cumulative learning progression reward (line \ref{alg:gentscl_learningprogression}). For the \textit{teacher}'s $\textsc{UpdateRule}$, we focus on multi-armed bandit learning (see~\autoref{sec:background-bandits}).
We adopt a data-centric perspective to perform a systematic investigation of its components.
% \end{figure}

\section{The Cooperative Mechanics of Experience} 
% \todo{TU disappeared from our discussion}
\label{sec:cooperative-game}

The ideal teacher-student interaction mechanics assume that the \textit{learner} monotonically increases its performance on the target task.
We conjecture that a prerequisite for this idealistic curriculum learning dynamics~\citep{matiisen2020curriculumbandit} to occur within~\ac{TSCL}-style algorithms is that experience (or data) presented to the \textit{learner} should not interfere with each other. 
In other words, \textit{units of experience} should interact cooperatively. 
From a game-theoretic perspective, we explain how these cooperative mechanics may emerge among units by examining the history of teacher-student interactions, the reward function, and the bandit selection policy.

\subsection{The Mechanics of Coalition Formation}

We establish a cooperative game where each unit of experience $\textbf{u} \in \textbf{U}$ is a {player}. Next, we interpret the history of $k\leq K$ teacher-student interactions $\textbf{H}_k =\{\textbf{u}_1, \ldots, \textbf{u}_{k}\}$ through their empirical frequencies $p_k(\textbf{u}) \in \Delta_\textbf{U}$ which form unit vectors that lie in the $|\textbf{U}|$-probability simplex $\Delta(\textbf{U})$. The effective support (i.e., non-zero probabilities) determines an unordered coalition (i.e., a set) $\textbf{C}_k \subseteq \textbf{U}$ (see~\citet{faigle2022gametheory}, Chapter 8), formed by the units presented to the \textit{learner} up to interaction $k \leq K$. 
We study this interpretation through
a cooperative game in characteristic function form (\autoref{sec:background-games}). 
% The \textit{characteristic function} $v: 2^\textbf{U} \to \mathbb{R}$ measures for every coalition $\textbf{C}_k \in 2^\textbf{U}$ what units have gained by acting jointly.

\begin{example}{\textbf{(Example 3.1 cont'd)}}
    In the \textit{class-as-unit} equivalence on \textsc{MNIST}, an {unordered} training coalition, e.g., the two-unit coalition $\textbf{C}=\{\textsc{zero}, \textsc{nine}\}$, describes teacher-student interactions limited to instances from those \textit{classes}. 
\end{example}

Next, we note that the outcome of a coalition's work is the policy or model $\pi_k$.  Thus, estimating the performance of the policy $\pi_k$ through the metric function $\mathcal{J}$ is akin to approximating the {characteristic function} $v(\textbf{C}_k)$ (Alg.~\ref{alg:gentscl}, line ~\ref{alg:gentscl_learningprogression}).
% In the particular case when no unit has been presented to the learner algorithm, the value of $\mathcal{J}(\pi_0)$ estimates the characteristic function of the \textit{empty coalition}. 
% Thus, we consider the empty coalition  $\textbf{C}_0 = \varnothing$ and its worth $v(\textbf{C}_0 = \varnothing)$ a representation of the \textit{marginal effect of initialization}. 
% \footnote{Large values at initialization could indicate that no learning (i.e., and no curriculum) is required~\citep{oller2020randomguess}, or that the contribution of units may be of a smaller scale.}.
% \label{sec:cooperative-gamespace}
Moreover,
% as~\autoref{alg:gentscl} describes,
these approximations are conditioned on an evaluation (or target) unit $\bar{\textbf{u}} \in \bar{\textbf{U}}$. 
We model the \textit{target-task} and \textit{multiple-task} settings \citep{graves2017curriculumbandit} where \textit{units of experience} should increase \textit{learner} performance on an evaluation unit (e.g., a task, or an environment) or on multiple evaluation units (e.g., a set of tasks or environments). Consequently, every notion of coalitional worth is conditional on the evaluation units, thus generating a space of cooperative games. 

\begin{definition}{(\textbf{\ac{TSCL} Cooperative Games})}
Let $\textbf{U}$ denote a set of \textit{units of experience} $\textbf{u} \in \textbf{U}$ and $\bar{\textbf{U}}$ a set of \textit{evaluation units} $\bar{\textbf{u}} \in \bar{\textbf{U}}$.
Every evaluation coalition $\bar{\textbf{C}} \in 2^{\bar{\textbf{U}}}$
induces a parameterized characteristic function
$v_{\bar{\textbf{C}}}(\textbf{C}_k) \in \mathbb{R}$ whose value measures the worth 
of a coalition $\textbf{C}_k$ when the members of $\bar{\textbf{C}}$ are the \textit{evaluation units}. 
Therefore, the \ac{TSCL}-family of algorithms operate over a  parameterized space of cooperative games: $$\boldsymbol{\mathcal{G}}\left<\textbf{U}, \cdot\right> = \left\{\left<\textbf{U}, v_{\bar{\textbf{C}}} \right>~|~\bar{\textbf{C}} \subseteq \bar{\textbf{U}} \right\}$$ 
comprising $2^{|\textbf{U}|} \times 2^{|\bar{\textbf{U}}|}$ possible games and where the \textit{target-task} (i.e., $\bar{\textbf{C}}=\bar{\textbf{u}}$) and the \textit{multiple-tasks} (i.e., $\bar{\textbf{C}}=\bar{\textbf{U}}$) settings are special cases.
\end{definition}

\begin{example}
    If a \textit{learner} algorithm is presented with units from $\textbf{C}=\{\textsc{zero}, \textsc{nine}\}$ on \textsc{MNIST}, the following condition is expected to hold:
    $$
        v_{\bar{\textbf{C}}=\{\textsc{zero},  \textsc{nine}\}}(\textbf{C}) > v_{\bar{\textbf{C}}=\{\textsc{zero}, \textsc{one}\}}(\textbf{C})
    $$
\end{example}

% \subsection{Marginal Contributions to Learning}
% Fair attribution mechanisms allocate a value to players that is proportional to the increase in coalitional worth caused by them joining a coalition (i.e., their \textit{marginal contributions}). 
\subsection{Marginal Contributions to Learning}
\label{sec:marginal-contribution}

The notions of coalitions and coalitional worth above induce a game-theoretic interpretation of the learning progression reward. 
At any iteration $k \leq K$, this reward $r(\textbf{u}_k) \in \mathbb{R}$ (Alg.~\ref{alg:gentscl}, line~\ref{alg:gentscl_learningprogression}) measures the improvement in policy performance after the \textit{teacher} presents a unit $\textbf{u}_k$ to the \textit{learner} algorithm that produces a new policy $\pi_k \sim \mathcal{L}(\pi_{k-1}, \textbf{u}_k)$.
Thus, we can restate this reward in terms of a game in characteristic function form:
\begin{equation}
    r(\textbf{u}_k) = v(\textbf{C}_{k}) - v(\textbf{C}_{k-1}) = v(\textbf{C}_{k-1} + \textbf{u}_k) - v(\textbf{C}_{k-1}) 
    % = \nabla_{\textbf{u}_k} v(\textbf{C}_k) 
    \label{eq:learning-progression-coal}
\end{equation}
and note its equivalence to computing the marginal contribution (see~\autoref{sec:background} and~\autoref{eq:shapley}) of aggregating the unit of experience $\textbf{u}_k = \textbf{u}$ to the existing coalition $\textbf{C}_{k-1}$.
% Therefore, ~\autoref{eq:learning-progression-coal} provides an alternative interpretation to this intrinsic motivation objective~\citep{Schmidhuber_1991, Oudeyer_Kaplan_2009}.

\subsection{A Fair Allocation Mechanism}
\label{sec:fair-allocation}

A principle of fair attribution in cooperative games is that players get assigned values proportional to their expected marginal contribution.
We note that under the learning progression objective, a bandit action-value estimate $q_k(\textbf{u})$ (\autoref{eq:empirical_values}) approximates every unit's (or arm's) average marginal contribution after $k$ interactions. 
Moreover, as discussed in~\autoref{sec:background-bandits}, multi-arm bandit algorithms may transform action-values through a Boltzmann projection (\autoref{eq:boltzmann})  that converts the value estimates into units' probabilities of interaction with the \textit{learner} (i.e., the (stochastic) policy $\tau_k(\textbf{u})$). 
Consequently, the units that, up to interaction $k \leq K$, have produced more significant increases on \textit{learner} performance and would be allocated larger fractions of the remaining $K- k$ interactions. 

Thus, a {multi-armed bandit teacher} implements a {fair allocation mechanism} that computes {units' values} by approximating their {average marginal contributions}
 and converts these approximations into the currency-like utility of the \ac{TSCL} games, namely, {interactions with the learner}.  

\autoref{tab:equivalence} summarizes the equivalences between \ac{TSCL} components and those of a cooperative game we have established throughout this section.

\begin{table}[t]
    \centering
    \begin{tabular}{l|l|c|c}
        \toprule
         \textbf{TSCL} & \textbf{Cooperative Game} & \textbf{Learning} & \textbf{Example} \\
         \midrule
         Units &  Players &  & $\textbf{u}_1, \textbf{u}_2, \textbf{u}_3$ \\
         Sequence & Coalition & & $(\textbf{u}_2, \textbf{u}_1)$\\
         Policy Value & Coalitional Worth & $\mathcal{J}(\pi_2)$ & $v(\textbf{u}_2, \textbf{u}_1)$\\
         Learning Progression & Marginal Contribution & $\mathcal{J}(\pi_2) - \mathcal{J}(\pi_1)$ & $v(\textbf{u}_2, \textbf{u}_1) - v(\textbf{u}_2)$\\
         Unit Value & Player Value & $q_k(\textbf{u}_1)$ & $\phi(\textbf{u}_1)$\\
         Interactions & Allocations & $\tau_k(\textbf{u})$ & \\
         \bottomrule
    \end{tabular}
    \caption{The mechanics of Teacher-Student Curriculum Learning are equivalent to a cooperative game among units of experience.}
    \label{tab:equivalence}
\end{table}

\section{An Experiment on The Prospect of Cooperation}
\label{sec:expensive-prior}
% \todo{We trade exploration by enumeration. An a priori, fair allocation of the K interaction budget.}
% \todo{Curriculum among units exists even if TSCL fails to find it. Influence of the units sets on TSCL.}
% Among the many prospects game theory assumes are available to a player is the \textit{prospect of having to play a game}~\citep{shapley1952value}. 
We design an experimental setting to empirically verify the equivalences we draw between TSCL components and cooperative game-theoretic concepts (e.g., do cooperative solution concepts capture some notion of curriculum?) and to highlight the utility of this data-centric approach to understanding TSCL failure modes.
To this end, we incrementally build different parts of the equivalence in supervised learning, reinforcement learning, and classical game settings as follows: 
\begin{enumerate}
    \item For any set of \textit{units of experience} given to the \textit{teacher} algorithm, we simulate their interactions and compute each \textit{a priori} \textit{Shapley} or \textit{Nowak \& Radzik} value (\autoref{sec:simulation-cooperation}).
    \item We build two value-proportional mechanisms (i.e., pre-computed teacher policies) leveraging the prior units' values to validate whether a curriculum exists and to show that cooperative solution concepts retrieve such a notion.
    \item We leverage units' estimated pairwise interactions to understand the value-proportional mechanism success and TSCL failure from a data-centric perspective (i.e., the composition of the set of units).
\end{enumerate}

\subsection{A Simulation of Cooperation}
\label{sec:simulation-cooperation}

We simulate two coalition formation processes where \textit{units of experience} (e.g., classes, environments, or opponents) in each coalition {fairly} share a finite interaction budget $K \in \mathbb{N}$ and connect the value of the resulting learner's policies with coalitional's worth.

\subsubsection{Coalitional Mechanics and Worth}

\textbf{Cooperative (Non-Ordered) Game.} To simulate a {traditional cooperative game} (\autoref{sec:background}), we design a coalition formation process that draws at each interaction $k \leq K$ a unit $\textbf{u}_k \in \textbf{C}$ with uniform probability $\tau_\textbf{C}(\textbf{u}_k) \propto {|\textbf{C}|^{-1}}$, from a coalition $\textbf{C}$, 
and present it to a \textit{learner} algorithm. 
We compute this procedure for every {coalition} of units $\textbf{C} \in 2^{\textbf{U}}$.
The uniform distribution reflects an \textit{a priori} ignorance of units' importance before measuring their effect on the \textit{learner}.

\textbf{Generalized (Ordered) Game.} To test whether our formulation captures order for curriculum, we build a coalition formation process that simulates a {generalized cooperative game}. In this setting,  a \textit{unit}
$\textbf{u} \in \textbf{C}$ is continually presented to the \textit{learner}, 
for $\lfloor {K}/{|\textbf{C}|} \rfloor$ interactions, in its  permutation order
on an {ordered coalition} $\textbf{C}$. We repeat this procedure for every
$\textbf{C} \in \mathcal{P}(2^{\textbf{U}})$. As before, the ordered equipartition of interactions reflects our ignorance about units \textit{a priori} effect.

\textbf{Coalitional Worth \& Characteristic Function.} 
For every coalition, we obtain a model $\pi^K_{\textbf{C}}$ after $K$ interactions with the units in $\textbf{C}$ through either the traditional or generalized mechanics described above. Therefore, by the equivalence we established between policy performance and (conditional) coalitional worth (\autoref{sec:marginal-contribution}), the evaluation of each policy determines the characteristic function $v(\textbf{C})$. More importantly, because the resulting policy $\pi^K_{\textbf{C}}$ is unbiased for the \textit{evaluation units or coalitions} (e.g., the uninformed priors in the mechanics), we estimate, using the same policy $\pi^K_{\textbf{C}}$, the coalitions worth for every {characteristic function} $v_{\bar{\textbf{C}}}(\textbf{C})$ parameterized by every evaluation or target coalition $\bar{\textbf{C}}$.

% By estimating every coalition's worth, we have the complete specification of a cooperative game. 

% The resulting policy or model is {not biased} for any \textit{evaluation unit or coalition}. 

\begin{example}{\textbf{(Example 4.1 cont'd)}} 
    Assume a budget of $K = 100$ interactions and a subset (coalition) of classes from ~\textsc{MNIST}, for instance, units (classes) $\textsc{zero}$ and $\textsc{nine}$. In the simulation of a traditional game, for a coalition $\textbf{C} = \{\textsc{zero}, \textsc{nine}\}$, we uniformly draw instances from each unit with probability $\tau(\textbf{u}) = \frac{1}{2}$. For a coalition $\textbf{C}=[\textsc{zero}, \textsc{nine}]$ in a {generalized game}, instances from unit $\textsc{zero}$ are presented for the first $k=50$ iterations followed by $k=50$ instances from $\textsc{nine}$. The resulting policy $\pi^K_{\textbf{C}}$ performance is equivalent to the value of the characteristic function $v(\textbf{C})$, evaluated at coalition $\textbf{C}$.
\end{example}

% \subsubsection{Solution Concepts}
\textbf{Marginal Contributions \& Solution Concepts.} 
The players {marginal contributions} are the central quantity of cooperative solution concepts. What do marginal contributions capture in our experiments? First, in both coalition formation processes, adding a unit $\textbf{u}$ to a coalition $\textbf{C}$ while keeping the number of interactions $K$ constant reduces the learner algorithm's interactions with the existing units. 
For instance, in the {traditional cooperative game} simulation, adding a unit $\textbf{u}^\prime$ reduces the probability of drawing any unit already in $\textbf{u} \in \textbf{C}$ from $p_{\textbf{C}}(\textbf{u})=|\textbf{C}|^{-1}$ to $p_{\textbf{C} + \textbf{u}^\prime}(\textbf{u})=(|\textbf{C}|+1)^{-1}$, while in the the {generalized game} reduces units' interactions from $\lfloor K / |\textbf{C}| \rfloor$ to $\lfloor K / (|\textbf{C}|+1) \rfloor$.
Therefore, in either simulation, 
a unit $\textbf{u}$ {marginal contribution} $v(\textbf{C} + \textbf{u}) - v(\textbf{C})$ measures the change in performance produced by increasing learner's interactions with $\textbf{u}$ while reducing those with the existing units in $\textbf{C}$.

\begin{example}{\textbf{(Example 5.1 cont'd)}}
     In a {traditional cooperative game simulation} on \textsc{MNIST}, a marginal contribution such as $v(\{\textsc{zero}, \textsc{nine}\}) - v(\{\textsc{zero}\})$ measures the change in \textit{learner} performance produced by exchanging approximately $K/2$ interactions with unit $\textsc{zero}$ for interactions with $\textsc{nine}$. However, in the {generalized game simulation}, the same expression measures the change in performance produced by exchanging $K$ interactions with unit $\{\textsc{zero}\}$ for $K/2$ with $\textsc{zero}$ first (pre-training) followed by $K/2$ with $\textsc{nine}$ (fine-tuning).   
\end{example}
In consequence, both solution concepts leveraging marginal contributions, namely, the \textit{Shapley value} for {traditional games} (\autoref{eq:shapley}) and the \emph{Nowak \& Razik's value} for {generalized games} (\autoref{eq:nowak_radzik}) estimate, in our simulations, a unit's average {marginal contribution} to learning, and thus capture their {helpfulness} or {cooperativeness} for curriculum.

\subsubsection{A Sanity Check Through Supervised Classification}
\label{sec:supervised-clasification}
Our running examples on~\textsc{MNIST} inspire the first setting we examine to empirically validate the connections we have established between TSCL and cooperative games. In this experiment, we considered instances aggregated in classes as \textit{units of experience} and benefited from a trained classifier's confusion matrix to extract ground-truth information from unit interactions. 
For both~\textsc{MNIST} and~\textsc{CIFAR10}~\citep{krizhevsky2009cifardatesets}, we trained a model on the complete dataset (e.g., for $200$ epochs), extracted the confusion matrix on validation, and identify the \textit{top-k} most confused pairs of classes. 
On \textsc{MNIST}, we selected the five classes  $\textsc{two}, \textsc{three}, \textsc{four}, \textsc{five}$ and $\textsc{seven}$ belonging to the top-three most confused pairs (see~\autoref{sec:prospect-prior-hparams}, \autoref{fig:mnist-truth}), grouped their instances into {five} \textit{units of experience}, and conduct the approximations described in ~\autoref{sec:simulation-cooperation} for a traditional cooperative game (i.e., not considering order). Next,
we followed the same approach for~\textsc{CIFAR10} six classes with more significant pairwise confusion errors on validation, namely, \textsc{car}, \textsc{cat}, \textsc{deer}, \textsc{dog}, \textsc{frog} and \textsc{horse} (see~\autoref{sec:prospect-prior-hparams}, \autoref{fig:cifar10-truth}). 

If the equivalences we drew are correct, one may expect a unit Shapley value to be higher when the evaluation target is the same unit. Moreover, the pairwise interactions between units should approximately reflect the negative interactions we extracted from the confusion matrices of each trained classifier.

\begin{figure}[t]
    \centering
    % \advance\leftskip-0.7cm
    \begin{subfigure}{0.48\textwidth}
        \begin{subfigure}{0.48\textwidth}
            \centering
            \includegraphics[width=\textwidth, height=2.55cm]{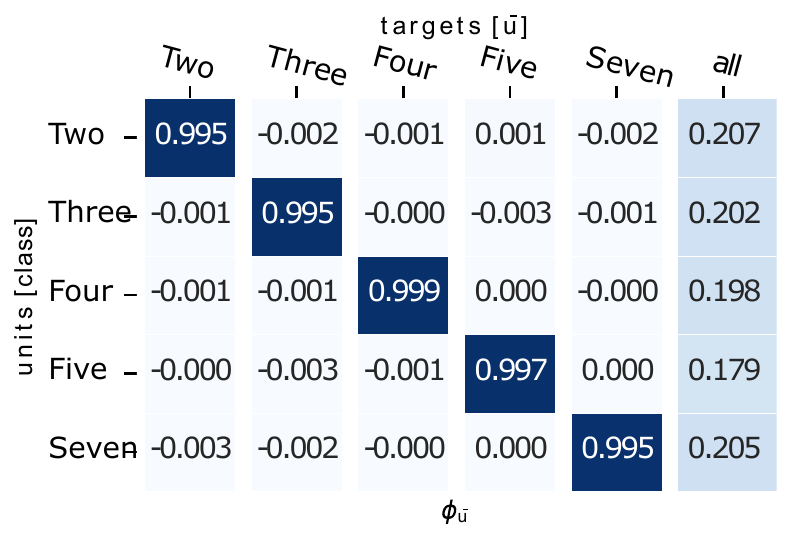}
            \caption{}
            \label{fig:MNIST-shapley}
        \end{subfigure}
        \begin{subfigure}{0.48\textwidth}
            \centering
            \includegraphics[width=\textwidth, height=2.55cm]{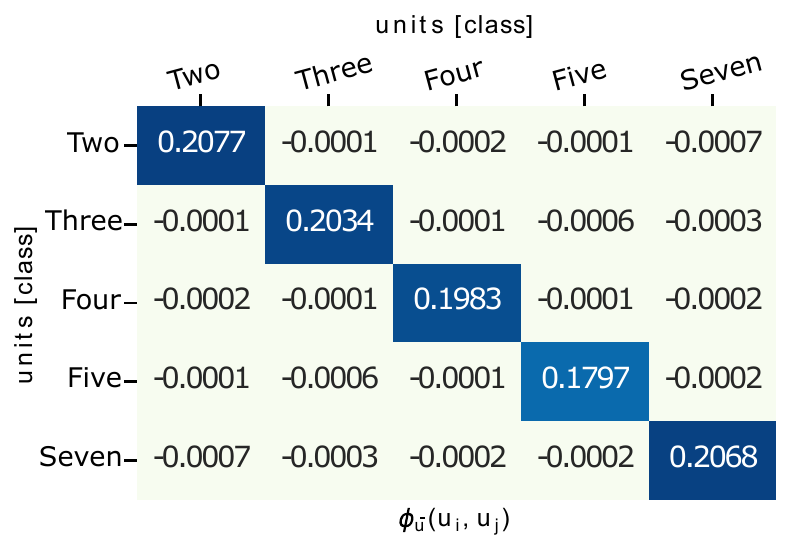}
            \caption{}
            \label{fig:MNIST-vpop}
        \end{subfigure}
        % \caption*{\tiny \textsc{MNIST}}
    \end{subfigure}
    \begin{subfigure}{0.48\textwidth}
        \begin{subfigure}{0.48\textwidth}
            \centering
            \includegraphics[width=\textwidth, height=2.5cm]{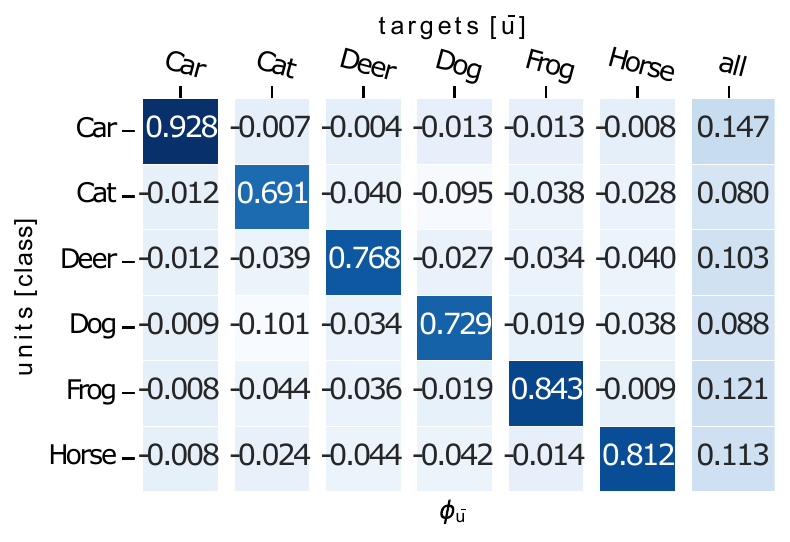}
            \caption{}
            \label{fig:CIFAR-shapley}
        \end{subfigure}
        \begin{subfigure}{0.48\textwidth}
            \centering
             \includegraphics[width=\textwidth, height=2.5cm]{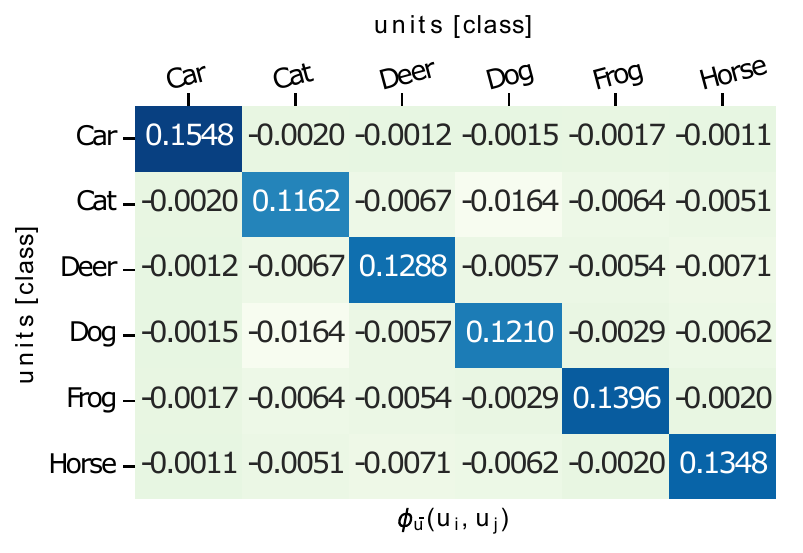}
             \caption{}
             \label{fig:CIFAR-vpop}
        \end{subfigure}
        % \caption*{\tiny \textsc{Cifar10}}
    \end{subfigure}
    
    \caption{\small We validated the \textbf{prospect prior} using the \textit{class-as-a-unit} analogy on \textsc{MNIST} and \textsc{Cifar10}.
    In Figures (a) and (c), each column represents units' Shapley values $\phi(\textbf{u})$ in each cooperative game parameterized by a target-unit $\bar{\textbf{u}}$ and the target coalition of \textit{all} units. In Figures (b) and (d), we present the \ac{vPoP} decomposition matrix (\autoref{eq:vpop}) measuring the pairwise interaction values $\phi(\textbf{u}_i, \textbf{u}_j)$ among units in the \emph{all-units} target.
    % Units individual Shapley values produce correct valuations in each \emph{target-unit} (self-importance) and \emph{all-units} (equal-importance) settings (a, c).  When the target is \textit{all} classes, \ac{vPoP} identifies \textit{interfering} pairs, matching the ground truth confusion matrices (\autoref{sec:prospect-prior-hparams}). For instance, digits $2$ \& $7$ on \textsc{MNIST} and classes \textit{dog} \& \textit{cat} on \textsc{Cifar10}.
    }
    \label{fig:validation}
\end{figure}

\textbf{Units' Values.} In effect,
for either \textsc{Mnist} or \textsc{Cifar10}, each unit's Shapley value estimated from the {traditional cooperative game} simulations correctly matches the ground truth information. In the {target-unit} setting, where each \textit{unit of experience} is also used as {evaluation unit},
each \textit{unit of experience} (or class) matching the {evaluation unit} (or class) has the largest \textit{Shapley value}, as depicted in~\autoref{fig:MNIST-shapley} (first \textit{five} targets) for~\textsc{MNIST} and~\autoref{fig:CIFAR-shapley} (first \textit{six} targets) for~\textsc{CIFAR10}.
For instance, on \textsc{MNIST}, unit $\textbf{u} = \textsc{two}$ has the largest \textit{Shapley value} $\phi(\textsc{two})=0.995$ when the evaluation unit is $\bar{\textbf{u}}=\textsc{two}$. We observed a similar effect on \textsc{Cifar10}.
Furthermore, for the \textit{all-units} setting, every unit's \textit{Shapley value} is approximately equal on both \textsc{MNIST} and \textsc{CIFAR10} (\autoref{fig:MNIST-shapley} and \ref{fig:CIFAR-shapley}, \textit{all} column), matching the intuition that, conditional on an \textit{all} units evaluation, every \textit{unit of experience} should be equally valuable. 
% These results confirm the \textit{prospect prior}'s ability to estimate units' values correctly.

\textbf{Measures of Interactions.} 
We also computed the \ac{vPoP} measure (see~\autoref{sec:background-games},  \autoref{eq:vpop}) to verify whether its decomposition of \textit{Shapley values} into pairwise interaction values correctly identifies the most confused pairs of classes. 
For both~\textsc{MNIST} and ~\textsc{CIFAR10}, their respective \ac{vPoP} matrices, displayed in~\autoref{fig:MNIST-vpop} and~\ref{fig:CIFAR-vpop}, provide a reasonable approximation to the ground-truth pairwise interactions extracted from the confusion matrices.
For instance, on \textsc{MNIST} the \textit{units} $\textsc{two}$ and $\textsc{seven}$ have the lowest interaction value $\phi(\textsc{two}, \textsc{seven}) = \phi(\textsc{seven}, \textsc{two}) = -0.007$ which corresponds to largest entry  $M(2, 7) = 20$ of the confusion matrix (see~\autoref{sec:prospect-prior-hparams}, \autoref{fig:mnist-truth}). 
Also, in \textsc{Cifar10} the units \textsc{dog} and \textsc{cat} have the lowest interaction value $\phi(\textsc{dog}, \textsc{cat}) = -0.0164$ coinciding with the most confused classes $M(\textsc{dog}, \textsc{cat}) = 66$ on validation (see~\autoref{sec:prospect-prior-hparams}, \autoref{fig:cifar10-truth}). However, we note that, for instance, on~\textsc{MNIST} confusion matrix $M(2, 7) \neq M(7,2)$ and, similarly, on \textsc{CIFAR10}'s $M(\textsc{dog}, \textsc{cat}) \neq M(\textsc{cat}, \textsc{dog})$.
Nevertheless, we interpret these values as reasonable proxies for units \emph{negative, positive, or neutral} pairwise interactions.

% \subsubsection{Other Learning Paradigms}
A valuable aspect of our cooperative game-theoretic analysis of TSCL is that the equivalences we introduced are not limited to the supervised classification setting where ground truth information is available. Next, we demonstrate the broad applicability of these ideas through problems in \ac{RL} and classical games, where finding ground-truth notions of \textit{curriculum} is non-trivial. Details to reproduce these experiments are provided in \autoref{sec:prospect-prior-hparams}.

\subsubsection{Reinforcement Learning}  
We investigate the \textsc{Minigrid-Rooms}~\citep{minigrid} set of three environments, namely, \textsc{TwoRooms}, \textsc{FourRooms}, and \textsc{SixRooms} for which it is \textit{folk knowledge} that an optimal curriculum exists.\footnote{The curriculum order between these three environments in RL is an intuition that, to our knowledge, has not been quantified before.} 
We apply the \emph{prospect prior} simulation of a {generalized game} where we consider each {environment} a \emph{unit of experience}. 
As a \textit{learner} algorithm, we used \textsc{PPO}~\citep{schulman2017proximal} with an interaction budget of $K=500,000$ steps, and estimated, from the outcome of these simulations, 
the \textit{Nowak \& Radzik} values (\autoref{sec:background}, \autoref{eq:nowak_radzik}), with every environment and a uniform distribution over \textit{all}, as evaluation targets. 
In~\autoref{fig:minigrid-nr}, we show that the \emph{Nowak \& Radzik} values we estimate match \emph{folk knowledge}. First, there is no requirement for environments other than $\textsc{TwoRooms}$ as the only positive value $\phi_{\bar{\textbf{u}}}(\textsc{TwoRooms})=0.423$ correctly measures. Then, for $\textsc{FourRooms}$,  the values of $\phi_{\bar{\textbf{u}}}(\textsc{TwoRooms})=0.041$ and  $\phi_{\bar{\textbf{u}}}(\textsc{FourRooms})=0.107$ indicate that both environments are required. And finally, environments values of $\phi_{\bar{\textbf{u}}}(\textsc{FourRooms})=0.03$ and $\phi_{\bar{\textbf{u}}}(\textsc{SixRooms})=0.03$ 
indicate that both are needed for solving \textsc{SixRooms}.

\subsubsection{Classical Games}
We introduce an experimental setting, the \ac{A-SIPD}, that utilizes the {Prisioner's Dilemma}~\citep{merrillflood1952pd, axelrod_hamilton1981cooperation} classical two-player game as a base but in a more challenging \textit{sparse} and {iterated} version where at the end of a finite number of interactions (e.g., $200$ steps), a \textit{win-draw-loss} {reward} is given to the learner if it accrues more cumulative \textit{payoff} than its opponent. Opponents are drawn from a \textit{population} of five well-known strategies: \textsc{AlwaysCooperate}, \textsc{AlwaysDefect}, \textsc{WinStayLoseSwitch}, \textsc{TitForTat}~\citep{axelrod1981cooperation} and a \textsc{ZeroDeterminant} strategy
~\citep{ hilbe_nowak_sigmund_2013zerodet}.
We apply the \emph{prospect prior} simulation of a \textit{generalized game} where we consider each \textit{opponent} a \emph{unit of experience}. 
As a \textit{learner} algorithm, we used \textsc{PPO}~\citep{schulman2017proximal} with a budget of $K=100,000$ interactions. 

\begin{figure}[t]
    \centering
    \small
    \begin{subfigure}{0.48\textwidth}
        \begin{subfigure}{0.48\textwidth}
            \centering
            \includegraphics[width=\textwidth, height=2.5cm]{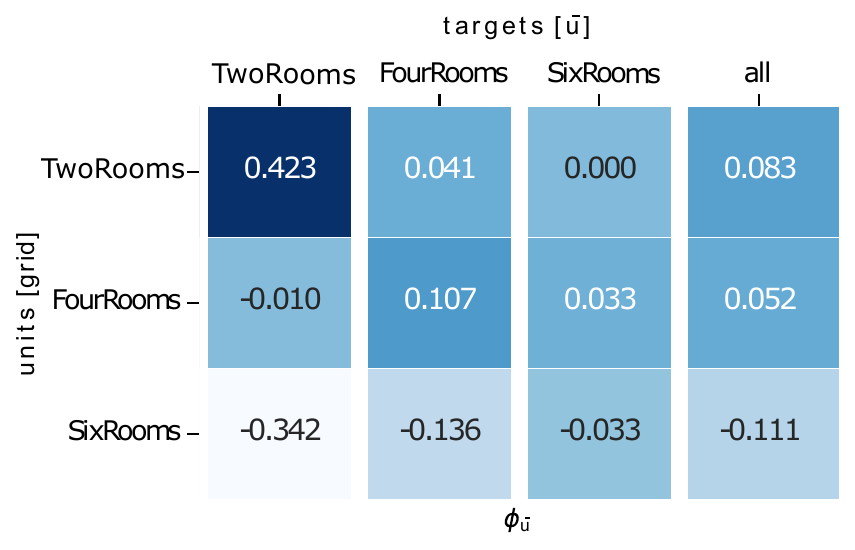}
            \caption{\textit{Nowak \& Radzik Values}}
            \label{fig:minigrid-nr}
        \end{subfigure}
        \begin{subfigure}{0.48\textwidth}
            \centering
            \includegraphics[width=\textwidth]{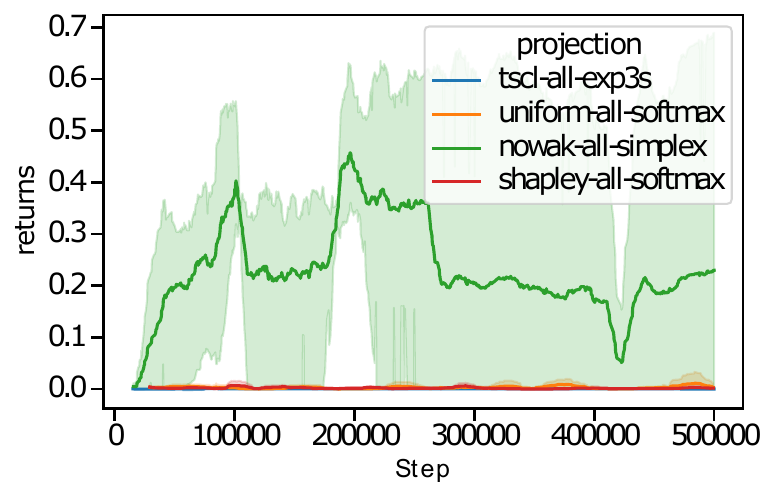} 
            \caption{Learning Curves [\textit{all}]}   
        \end{subfigure}
        \caption*{\textsc{MiniGrid-Rooms}}
    \end{subfigure}
    \begin{subfigure}{0.48\textwidth}
        \begin{subfigure}{0.48\textwidth}
            \centering
            \includegraphics[width=\textwidth, height=2.5cm]{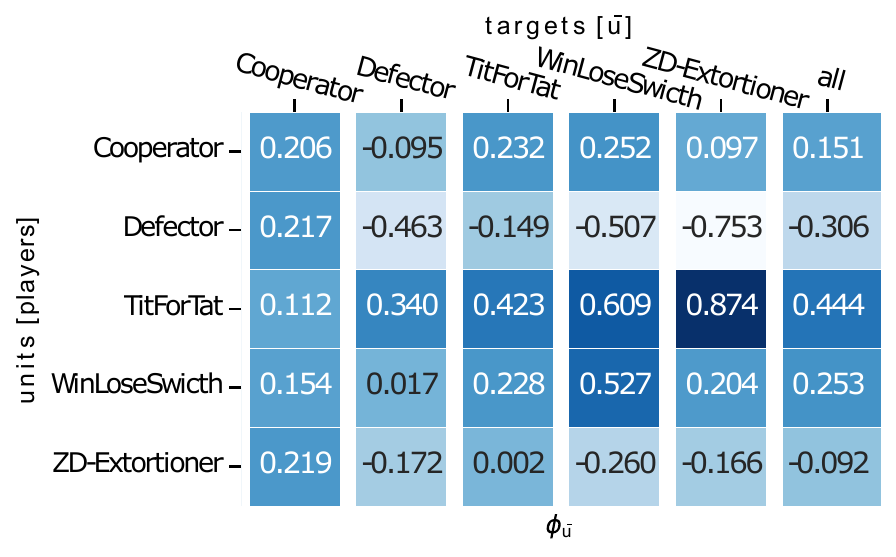}
            \caption{\textit{Nowak \& Radzik Values}}
            \label{fig:nowak-radzik-adv-sipd}
        \end{subfigure}
        \begin{subfigure}{0.48\textwidth}
                \centering
               \includegraphics[width=\textwidth]{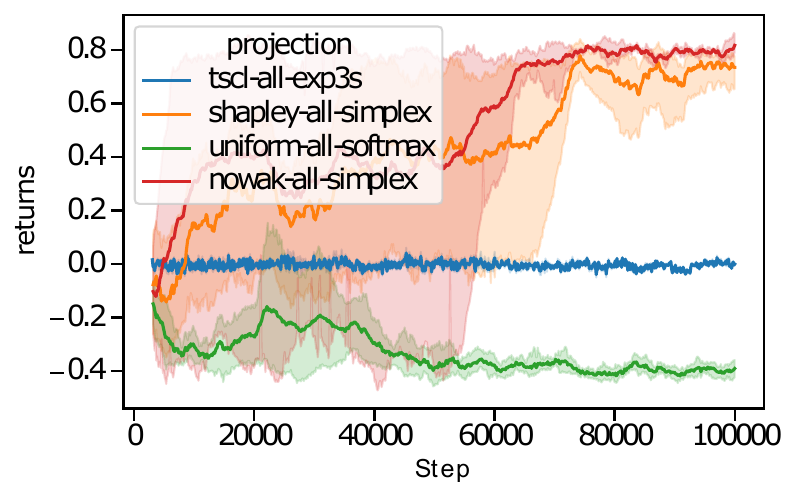}
            \caption{Learning Curves [\textit{all}]}   
        \end{subfigure}
        \caption*{\textsc{A-SIPD}}
    \end{subfigure}
    \caption{ \textit{Nowak \& Radzik} values \textit{(a, c)} conditional on each \emph{single-unit} and the \emph{multiple-units} (all) evaluations, and the \textit{learning curves} \textit{(b,d)} for our mechanisms and~\ac{TSCL}.
    }
    \label{fig:shapley-prospect}
\end{figure}

We estimated
the \textit{Nowak \& Radzik} values (\autoref{sec:background}, \autoref{eq:nowak_radzik}), conditioned on each opponent and a uniform mixture over \textit{all}, as evaluation targets. The results we present in~\autoref{fig:nowak-radzik-adv-sipd} show that playing uniquely against \textsc{TitForTat} is sufficient across all evaluation targets, including the most challenging opponents, \textsc{AlwaysDefect} and~\textsc{ZeroDeterminant}. This result contrasts with \textit{folk knowledge} in population-based training (e.g., playing against the population's \textit{Nash strategy}~\citep{nash1950equilibrium}). We defer to~\autoref{sec:tscl-population} a more in-depth discussion of this finding. 

\subsection{Curriculum from A Priori Values}
\label{sec:values-curriculum}

For both the RL and classical games settings, the Nowak \& Radzik value correctly \textit{ordered} tasks and opponents, respectively, by their contribution to learning. Nevertheless, we also investigated whether these values' magnitude indicated the proportion of interactions (i.e., a fraction of the $K$ interactions budget) that should be allocated to each unit and whether the combination of order and magnitude retrieved an approximate curriculum.

Therefore, inspired by {value-proportional allocations}~\citep{bachrach2020teamformation}, we
developed two mechanisms that turn units' values into interactions with the \textit{learner} by projecting any value vector $\phi(\textbf{u}) \in \mathbb{R}^{|\textbf{U}|}$ onto vectors $\tau(\textbf{u}) \in \Delta_\textbf{U}$ in a $|\textbf{U}|$-\textit{simplex}. 
While \citet{bachrach2020teamformation} use a \textit{linear} projection unable to handle negative marginal contributions, we investigate the \textit{Boltzmann} or \textit{softmax} projection,
commonly used in \textit{multi-arm bandit} algorithms (see~\autoref{sec:background-bandits}, \autoref{eq:boltzmann}), and an \textit{Euclidean} projection~\citep{blondel2014simplexproj} that projects to zero any unit with negative value. 

When values $\phi(\textbf{u})$ are \textit{Shapley values}, the projected vectors are used as pre-computed \textit{teacher} policies (i.e., probability distributions), mimicking ~\ac{TSCL}'s interactions but fixed \textit{a-priori} knowledge of units' value. However, when values $\phi(\textbf{u})$ are \emph{Nowak \& Radzik}, vectors $\tau \in \Delta_\textbf{U}$ are used as \textit{ordered compositional vectors}~\citep{aitchison1982compositional} 
that represent ordered fractions of $K$ interactions. Thus, we construct a pre-computed \textit{teacher} policy that, first orders units by their \emph{Nowak \& Radzik values}, projects the ordered values onto $\tau \in \Delta_\textbf{U}^K$, and presents the \textit{i-th ranked unit} $\textbf{u}_i \in \textbf{U}$ to the \textit{learner} for the number of interactions indicated by $\tau_i \in \mathbb{N}$. 
This mechanism preserves the ordered values captured by \textit{Nowak \& Radzik}'s solution concept.

\autoref{fig:shapley-prospect} shows
that for both~\textsc{Minigrid-Rooms} and  \textsc{A-SIPD} the \textit{teacher} policies obtained from these~\textit{value-proportional mechanisms}, in particular the {Euclidean} projection of \textit{Nowak \& Radzik} values (\emph{nowak-all-simplex}), consistently produce {learner-induced policies} that solve the target tasks, and empirically validate our hypothesis that \textit{a priori} values, in particular the generalized Nowak \& Radzik value we introduced, retrieve both the order of and the proportions of interactions that should be allocated to each unit to produce curriculum.

\begin{figure}[t]
    \centering
    \small
        \begin{subfigure}{0.48\linewidth}
        \centering
            \begin{subfigure}{0.48\textwidth}
                \centering
                \includegraphics[width=\textwidth, height=2.25cm]{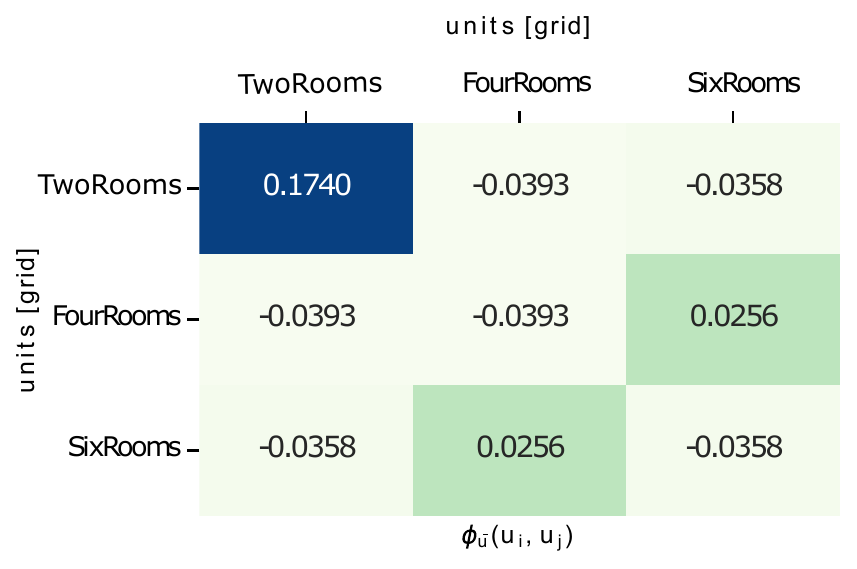}
                \caption{Shapley-vPoP}
            \end{subfigure}
            \begin{subfigure}{0.48\linewidth}
                \centering
                \includegraphics[width=\linewidth, height=2.25cm]{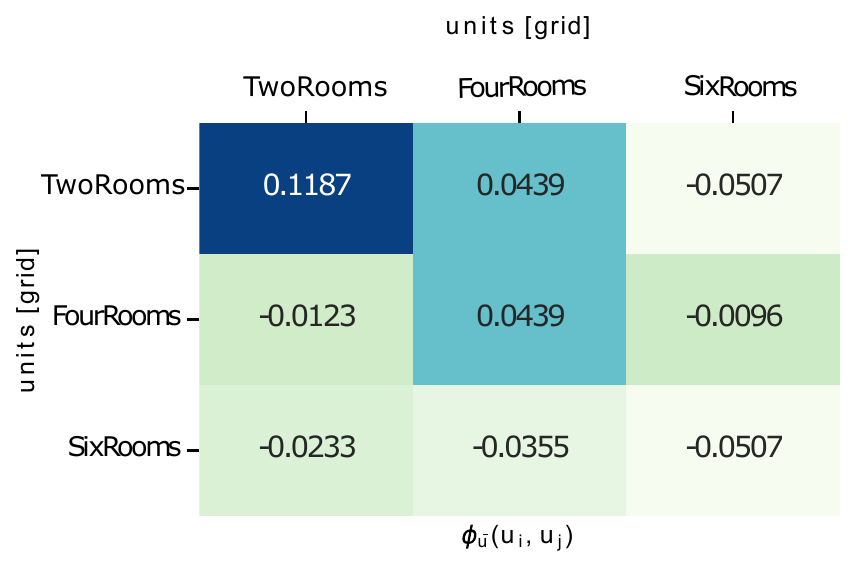}
                \caption{Nowak\&Radzik-vPoP}
            \end{subfigure}
        \end{subfigure}%
        \begin{subfigure}{0.48\linewidth}
        \centering
            \begin{subfigure}{0.48\linewidth}
                \centering
            \includegraphics[width=\linewidth, height=2.25cm]{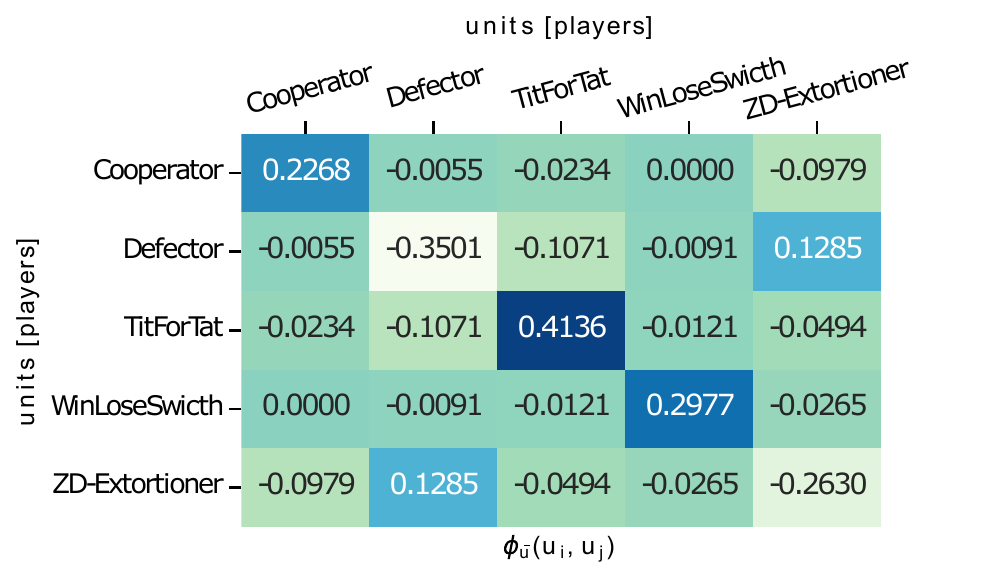}
                \caption{Shapley-vPoP}
            \end{subfigure}
            \begin{subfigure}{0.48\linewidth}
                \centering
                \includegraphics[width=\linewidth, height=2.25cm]{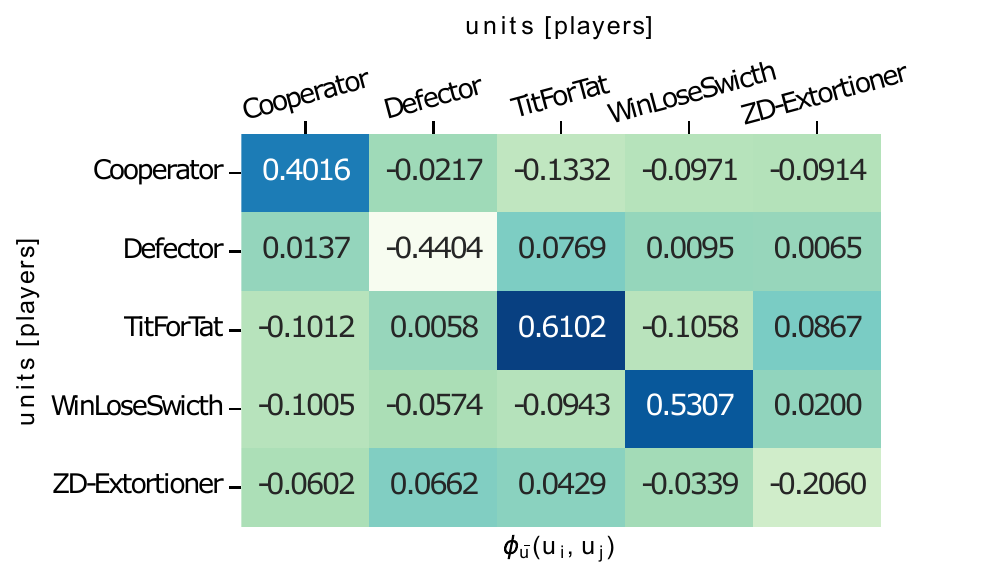}
                \caption{Nowak\&Radzik-vPoP}
            \end{subfigure}
        \end{subfigure}
    \caption{The~\ac{vPoP} decomposition of \textit{Shapley's} and \textit{Nowak \& Radzik values}, conditioned on the \textit{all-units} evaluation target, for the \textsc{MiniGrid-Rooms} \textit{(a, b)} and \textit{A-SIPD} \textit{(c, d)} problem settings.}
    \label{fig:vpop-analysis}
\end{figure}

\textbf{TSCL Data-Centric Failures.} Moreover,~\autoref{fig:shapley-prospect} also highlights the multi-arm bandit approach ($\textsc{Exp3}$~\citep{auer2003bandits, graves2017curriculumbandit}) to \ac{TSCL} (i.e., \emph{tscl-all-exp3s}) failure to produce an effective curriculum.
We found that in the presence of units with non-cooperative interactions, bandit-driven TSCL fails. 
\autoref{fig:vpop-analysis} presents the~\ac{vPoP} decomposition of \textit{Shapley} and \textit{Nowak \& Radzik} values conditioned on the \textit{all-units} evaluation.
In both settings, the interactions measured from the \textit{Shapley}-based decomposition produce lower (negative) values than those obtained with \textit{Nowak \& Radzik}. 
As we show through~\autoref{sec:cooperative-game} and~\autoref{sec:expensive-prior}, we take the \textit{Shapley}-based interaction values as fair approximations of~\ac{TSCL} interactions, and thus they
provide a data-centric explanation to~\ac{TSCL} failures (see~\autoref{sec:extended-expt-results}).
On the other hand, the \textit{Nowak \& Radzik}-based values explain the success of \textit{nowak-all-simplex} mechanism, along with the elimination, through the \textit{Euclidean} projection, of negatively-valued units. These pruning strategies are employed effectively by {data valuation} techniques~\citep{yanprocaccia2021shapleylovecore}.

% Moreover, there exist less challenging cases at which \ac{TSCL} excels, and we show some on~\textsc{Adv-SIPD} in \autoref{sec:extended-expt-results}. 

\subsection{Single and Population-based Curriculum}
\label{sec:tscl-population}

These results offer an alternative approach to what population-based training approaches prescribe as curriculum~\citep{lanctot2017unified, balduzzi2019openended, garnelo2021pick, liu2022neupl}.
Generally, \emph{meta-strategy solvers} for population-based training leverage tools from non-cooperative game theory~\citep{vonneumann1944theory} to find, for instance, the mixed \emph{Nash equilibrium}~\citep{nash1950equilibrium} of the \textit{empirical meta-game}~\citep{wellman2006egt} played by the population of opponents.
In this sense, we may also understand \ac{TSCL} as a \textit{cooperative meta-strategy solver} that prioritizes among a fixed population of (non-learning) opponents (units of experience) those that improve the learning progression of a single learning player against one or more opponents of the same population.

In the sparse and iterated version of Prisoner's Dilemma that we introduced, the \emph{Defector} strategy remains the (empirical) game \textit{Nash equilibrium}. 
However, the ordered prospect prior results in ~\autoref{fig:nowak-radzik-adv-sipd} show that when evaluated on the population Nash $\bar{\textbf{u}} = \textit{Defector}$, 
the largest \textit{Nowak \& Radzik} value corresponded to the \emph{TitForTat} strategy $\phi_{\bar{\textbf{u}}}(\textit{TitForTat})=0.34$. Playing against the \emph{TitForTat} strategy remains the optimal solution across all evaluation targets, meaning that \emph{TitForTat} is the best proxy opponent to learn from and reach the Nash equilibrium strategy.
We can interpret this result from two perspectives. First, it could indicate that the sparse, iterated, and overtly adversarial version of the game we constructed is a more complicated problem than the original, and the Nash equilibrium, the \emph{Defector} strategy, is a stronger opponent. However, these results may also indicate that a cooperative approach to meta-strategy selection may improve performance in some scenarios. We believe that this connection warrants further investigation.

\section{Limitations}
% We would not consider the \textit{prospect prior} an algorithmic contribution to~\ac{TSCL} but an analysis tool.
The simulations we computed in our experiments are computationally expensive. It is well-established that cooperative solution concepts are NP-hard~\citep{deng_papadimitriou1994complexitycoop, elkind2009solutionconceptscomputational}. However, better approximations are possible~\citep{yanprocaccia2021shapleylovecore, mitchell2022samplingshapley} although we do not explore them in this work. 
Consequently, we do not consider the \textit{prospect prior} experiments and the \textit{value-proportional curriculum mechanism} as algorithmic innovations to replace \ac{TSCL}. 
They represent a \textit{data-centric} approach to study the limits imposed on \ac{TSCL}-style algorithms by the (non)cooperative mechanics among units of experience. However, we acknowledge that the mechanics of units' interactions also affect other aspects of~\ac{TSCL}. These aspects may include, for instance, the \textit{teacher}'s credit-assignment problem~\citep{gittins1979bandits} or neural networks learning and forgetting dynamics (e.g., ~\citet{Lee_Goldt_Saxe_2021}). We control for these factors by keeping them constant in our experiments (see~\autoref{sec:prospect-prior-hparams}) but do not undertake their analysis here. 
Our work is a starting point for more thorough explorations of \ac{TSCL} and curriculum learning, their underlying mechanisms and broader applicability in machine learning.

\section{Related Work}
\label{sec:extended-related-work}

\paragraph{Curriculum Learning.} Since the seminal works of \citet{elman1993curriculum, krueger_dayan2009curriculum} and \citet{bengio2009curriculum},  a large body of literature has been produced on curricula for machine learning algorithms. Excellent surveys presented by ~\citet{narvekar2022surveyclrl, wang2022surveycl}, and \citet{soviany2022surveycl} offer a comprehensive state of recent advances in the field. Our work closely inspects the \ac{TSCL} framework concurrently introduced by~\citet{graves2017curriculumbandit} and~\citet{matiisen2020curriculumbandit}.  While follow-up works have generally focused on either algorithmic innovations \citep{weinshall2018curriculum, portelas2020metacurriculum, feng2021curriculumbandit, liu2020curriculumbanditadam, portelas2019teachercurriculum, Racaniere2020curriculumsettersolver,florensa2017reversecl,campero2021curriculum, du2022curriculumselfplay, florensa2018curriculumgan} or evaluation benchmarks \citep{babyai_iclr19, romac2021curriculum}, the \textit{data-centric perspective}~\citep{ng2021data, Karpathy2021tesla} we proposed here is less explored. The work by \citet{Wu2020-hq} empirically verifies the same fundamental questions on \emph{why}, \emph{when}, and \emph{how} curriculum learning works. However, while our work is more specifically focused on~\ac{TSCL}, we provide a game-theoretic grounding that could be further extended to analyze~\citet{Wu2020-hq} setting. 
On the other hand, \citet{Lee_Goldt_Saxe_2021} specifically explore~\ac{TSCL} but with a focus on the deep neural network dynamics leading to \textit{catastrophic forgetting}~\cite{mccloskey1989catastrophic}. The evaluation mechanisms that we propose in the \textit{prospect prior} simulation and the game-theoretic characterization of experience interference we introduce here provide a more general grounding for the problems studied there. 

\paragraph{Machine Learning \& Game Theory.} Our work draws much of its inspiration from the tradition of cross-pollination between machine learning and game theory research~\citep{cesa-bianchi_lugosi_2006}. In recent years, game theory research has fuelled work in \textit{optimization}~\citep{daskalakis2018optimgame, jin2019gameoptim}, \textit{generative modelling}~\citep{goodfellow2014gans, farnia2020gangames, mohebbi2023gangames}, or \textit{robustness}~\citep{madry2018adversarialattacks, huang2022adverarialrl}. Also, game-theoretic arguments have been used to revitalize long-standing problems in machine learning. For instance, \citet{gemp2021eigengame} reframed the century-old problem of \textit{Principal Component Analysis} (PCA)~\citep{pearson1901pca,hotelling1933pca} as the \textit{Nash} equilibrium~\citep{nash1950equilibrium} of a multiplayer game. Recently, 
\citet{chang2020decisionmakingeconomy} reformulated sequential decision-making problems (i.e., RL~\citep{sutton2018rlintro}) as local economic transactions among specialized self-interested agents. Moreover, and closer to our data-centric approach, \citet{lundberg2017shap} pioneered work on \textit{feature attribution} and \emph{explainable machine learning models} through cooperative game theory. In particular, they approximate Shapley's value~\citep{shapley1952value} (\autoref{eq:shapley}) to compute the influence \textit{input features} have on model \textit{predictions}. In a way, our work extends~\citet{lundberg2017shap} \textit{input-feature-as-a-player} analogy through the notion of \textit{units of experience} and introduces substantial elements to discuss how the cooperative mechanics of experience affects~\ac{TSCL} at different levels of abstraction (i.e., from features to tasks) and varied learning paradigms. 

\textbf{Active Learning.}  As~\citet{graves2017curriculumbandit} and~\citet{matiisen2020curriculumbandit} noted in their works, ~\ac{TSCL} is an experience-prioritization mechanism holding a solid connection to algorithms that actively select experience. The central problem in \textit{active learning} is to decide which experience to present to a learning algorithm by querying an expert or oracle~\citep{settles2010activelearning}. Areas of research that have also adopted this paradigm of \emph{actively sampling experience} include \emph{experience replay} mechanisms in \ac{RL}~\cite{ schaul2016per, andrychowicz2017her, fedus2020experiencereplay} where \textit{transitions} stored in a \emph{replay buffer}~\citep{lin1992experiencereplay} are sampled according to some prioritization mechanism. This category of active methods fits perfectly under the TU-game among units of experience we introduced in~\autoref{sec:cooperative-game}.

\paragraph{Multitask Learning.} The cooperative game we designed also provides grounding to the problem of determining \textit{which tasks} should be learned together and \textit{what their optimal order is}, originally from the literature of \textit{multitask learning}~\citep{caruana1997multitask}. The discussion on \textit{experience interference} in~\autoref{sec:cooperative-game}, the \emph{prospect prior} simulation and our computation of \textit{ordered} and \textit{unordered} Shapley values, and the \ac{vPoP} metric are some novel tools we introduce that could shed new light on this problem. Recent work has explored the different avenues of how \textit{task interference} and \textit{ordering} affect learning outcomes~\citep{Standley2019-sk, Shamsian2023-gf, Zhang2021-ip, Fifty2021-wi, Lin2019-ew}. Our work provides a novel grounding that relates easily to these efforts. 

\paragraph{Continual Learning.} Finally, another area of interest to our work is \textit{continual learning}~\cite{Parisi2019continuallearning, delange2022continuallearning, mundt2023continuallearning}. We note that the \textit{ordered prospect prior} simulates a \textit{continual learning} setting. Units of experience are presented to the learning algorithm in order, and their importance (i.e., value in the cooperative game) is estimated using \citet{Nowak1994shapleygen} extension (\autoref{eq:nowak_radzik}) to Shapley's value. This perspective offers a principled approach to rank and order tasks by their importance, a critical aspect of the continual learning setting.
We believe that future extensions to the ordered prospect prior experiment may be part of a toolbox to understand continual learning benchmarks~\citep{lomonaco17core50, lin2021clear, srinivasan2022climb}, although we do not undertake that task here.

\section{Conclusions \& Future Work}
We reexamined~\ac{TSCL} through the lens of cooperative game theory. By drawing inspiration from work on {data valuation}, {feature attribution} and {explainability}, we provide a novel {data-centric} perspective that re-frames several of its components through alternative cooperative game-theoretic interpretations.
Our experiments confirmed the appropriateness of studying~\ac{TSCL}-style under this framework and highlighted the impact of units' cooperative mechanics on this problem.
However, we only began to unveil the potential of allocation mechanisms, solution concepts, and measures of interactions to explain some fundamental aspects of~\ac{TSCL} and hope our work inspires an influx of novel game-theoretic approaches to the problem.  Future work would explore more theoretically-grounded analysis of this problem through the connection between convex games~\citep{shapley1971coreconvex} and super(sub)modularity in discrete combinatorial optimization~\citep{dughmi2009submodularity, bach2011submodularity, krause2011submodularity} and the extension to continuous set of units through values of non-atomic games~\citep{aumanshapley1974nonatomic}.

\section*{Acknowledgements}

MD would like to thank Eugene Vinitsky for insightful feedback on earlier versions of this manuscript. The NSERC Discovery Grant and the Canada CIFAR AI Chair program supported MD and LP. Computing resources provided by Mila, Qu\'ebec AI Institute, partly enabled this research.

% \newpage
\bibliography{main}

\begin{thebibliography}{111}
\providecommand{\natexlab}[1]{#1}
\providecommand{\url}[1]{\texttt{#1}}
\expandafter\ifx\csname urlstyle\endcsname\relax
  \providecommand{\doi}[1]{doi: #1}\else
  \providecommand{\doi}{doi: \begingroup \urlstyle{rm}\Url}\fi

\bibitem[Aitchison(1982)]{aitchison1982compositional}
J~Aitchison.
\newblock The statistical analysis of compositional data.
\newblock \emph{Journal of the Royal Statistical Society}, 44\penalty0 (2):\penalty0 139--160, January 1982.

\bibitem[Andrychowicz et~al.(2017)Andrychowicz, Wolski, Ray, Schneider, Fong, Welinder, McGrew, Tobin, Pieter~Abbeel, and Zaremba]{andrychowicz2017her}
Marcin Andrychowicz, Filip Wolski, Alex Ray, Jonas Schneider, Rachel Fong, Peter Welinder, Bob McGrew, Josh Tobin, Openai Pieter~Abbeel, and Wojciech Zaremba.
\newblock Hindsight experience replay.
\newblock In I~Guyon, U~Von Luxburg, S~Bengio, H~Wallach, R~Fergus, S~Vishwanathan, and R~Garnett (eds.), \emph{Advances in Neural Information Processing Systems}, volume~30. Curran Associates, Inc., 2017.

\bibitem[Auer et~al.(2003)Auer, Cesa-Bianchi, Freund, and Schapire]{auer2003bandits}
Peter Auer, Nicol\`{o} Cesa-Bianchi, Yoav Freund, and Robert~E. Schapire.
\newblock The nonstochastic multiarmed bandit problem.
\newblock \emph{SIAM J. Comput.}, 32\penalty0 (1):\penalty0 48–77, jan 2003.
\newblock ISSN 0097-5397.
\newblock \doi{10.1137/S0097539701398375}.
\newblock URL \url{https://doi.org/10.1137/S0097539701398375}.

\bibitem[Aumann \& Shapley(1974)Aumann and Shapley]{aumanshapley1974nonatomic}
R.~J. Aumann and L.~S. Shapley.
\newblock \emph{Values of Non-Atomic Games}.
\newblock Princeton University Press, 1974.
\newblock URL \url{http://www.jstor.org/stable/j.ctt13x149m}.

\bibitem[Axelrod \& Hamilton(1981)Axelrod and Hamilton]{axelrod_hamilton1981cooperation}
R~Axelrod and W~D Hamilton.
\newblock The evolution of cooperation.
\newblock \emph{Science}, 211\penalty0 (4489):\penalty0 1390--1396, March 1981.

\bibitem[Axelrod(1981)]{axelrod1981cooperation}
Robert Axelrod.
\newblock The emergence of cooperation among egoists.
\newblock \emph{The American political science review}, 75\penalty0 (2):\penalty0 306--318, 1981.

\bibitem[Bach(2011)]{bach2011submodularity}
Francis Bach.
\newblock Learning with submodular functions: A convex optimization perspective.
\newblock November 2011.

\bibitem[Bachrach et~al.(2020)Bachrach, Everett, Hughes, Lazaridou, Leibo, Lanctot, Johanson, Czarnecki, and Graepel]{bachrach2020teamformation}
Yoram Bachrach, Richard Everett, Edward Hughes, Angeliki Lazaridou, Joel~Z Leibo, Marc Lanctot, Michael Johanson, Wojciech~M Czarnecki, and Thore Graepel.
\newblock Negotiating team formation using deep reinforcement learning.
\newblock \emph{Artificial intelligence}, 288\penalty0 (103356):\penalty0 103356, November 2020.

\bibitem[Balduzzi et~al.(2019)Balduzzi, Garnelo, Bachrach, Czarnecki, Perolat, Jaderberg, and Graepel]{balduzzi2019openended}
David Balduzzi, Marta Garnelo, Yoram Bachrach, Wojciech Czarnecki, Julien Perolat, Max Jaderberg, and Thore Graepel.
\newblock Open-ended learning in symmetric zero-sum games.
\newblock In Kamalika Chaudhuri and Ruslan Salakhutdinov (eds.), \emph{Proceedings of the 36th International Conference on Machine Learning}, volume~97 of \emph{Proceedings of Machine Learning Research}, pp.\  434--443. PMLR, 2019.

\bibitem[Bengio et~al.(2009)Bengio, Louradour, Collobert, and Weston]{bengio2009curriculum}
Yoshua Bengio, J\'{e}r\^{o}me Louradour, Ronan Collobert, and Jason Weston.
\newblock Curriculum learning.
\newblock In \emph{Proceedings of the 26th Annual International Conference on Machine Learning}, ICML '09, pp.\  41–48, New York, NY, USA, 2009. Association for Computing Machinery.
\newblock ISBN 9781605585161.
\newblock URL \url{https://doi.org/10.1145/1553374.1553380}.

\bibitem[Besson(2018)]{SMPyBandits}
Lilian Besson.
\newblock {SMPyBandits: an Open-Source Research Framework for Single and Multi-Players Multi-Arms Bandits (MAB) Algorithms in Python}.
\newblock Online at: \url{github.com/SMPyBandits/SMPyBandits}, 2018.
\newblock URL \url{https://github.com/SMPyBandits/SMPyBandits/}.
\newblock Code at https://github.com/SMPyBandits/SMPyBandits/, documentation at https://smpybandits.github.io/.

\bibitem[Blondel et~al.(2014)Blondel, Fujino, and Ueda]{blondel2014simplexproj}
Mathieu Blondel, Akinori Fujino, and Naonori Ueda.
\newblock {Large-Scale} multiclass support vector machine training via euclidean projection onto the simplex.
\newblock In \emph{2014 22nd International Conference on Pattern Recognition}, pp.\  1289--1294, August 2014.

\bibitem[Campero et~al.(2021)Campero, Raileanu, Kuttler, Tenenbaum, Rockt{\"a}schel, and Grefenstette]{campero2021curriculum}
Andres Campero, Roberta Raileanu, Heinrich Kuttler, Joshua~B. Tenenbaum, Tim Rockt{\"a}schel, and Edward Grefenstette.
\newblock Learning with {\{}amig{\}}o: Adversarially motivated intrinsic goals.
\newblock In \emph{International Conference on Learning Representations}, 2021.
\newblock URL \url{https://openreview.net/forum?id=ETBc_MIMgoX}.

\bibitem[Caruana(1997)]{caruana1997multitask}
Rich Caruana.
\newblock Multitask learning.
\newblock \emph{Machine learning}, 28:\penalty0 41--75, 1997.

\bibitem[Cesa-Bianchi \& Lugosi(2006)Cesa-Bianchi and Lugosi]{cesa-bianchi_lugosi_2006}
Nicolo Cesa-Bianchi and Gabor Lugosi.
\newblock \emph{Prediction, Learning, and Games}.
\newblock Cambridge University Press, 2006.
\newblock \doi{10.1017/CBO9780511546921}.

\bibitem[Chang et~al.(2020)Chang, Kaushik, Weinberg, Griffiths, and Levine]{chang2020decisionmakingeconomy}
Michael Chang, Sid Kaushik, S~Matthew Weinberg, Tom Griffiths, and Sergey Levine.
\newblock Decentralized reinforcement learning: Global {Decision-Making} via local economic transactions.
\newblock In Hal~Daum{\'e} Iii and Aarti Singh (eds.), \emph{Proceedings of the 37th International Conference on Machine Learning}, volume 119 of \emph{Proceedings of Machine Learning Research}, pp.\  1437--1447. PMLR, 2020.

\bibitem[Chevalier-Boisvert et~al.(2018)Chevalier-Boisvert, Willems, and Pal]{minigrid}
Maxime Chevalier-Boisvert, Lucas Willems, and Suman Pal.
\newblock Minimalistic gridworld environment for gymnasium, 2018.
\newblock URL \url{https://github.com/Farama-Foundation/Minigrid}.

\bibitem[Chevalier-Boisvert et~al.(2019)Chevalier-Boisvert, Bahdanau, Lahlou, Willems, Saharia, Nguyen, and Bengio]{babyai_iclr19}
Maxime Chevalier-Boisvert, Dzmitry Bahdanau, Salem Lahlou, Lucas Willems, Chitwan Saharia, Thien~Huu Nguyen, and Yoshua Bengio.
\newblock Baby{AI}: First steps towards grounded language learning with a human in the loop.
\newblock In \emph{International Conference on Learning Representations}, 2019.
\newblock URL \url{https://openreview.net/forum?id=rJeXCo0cYX}.

\bibitem[Daskalakis \& Panageas(2018)Daskalakis and Panageas]{daskalakis2018optimgame}
Constantinos Daskalakis and Ioannis Panageas.
\newblock The limit points of (optimistic) gradient descent in min-max optimization.
\newblock In S.~Bengio, H.~Wallach, H.~Larochelle, K.~Grauman, N.~Cesa-Bianchi, and R.~Garnett (eds.), \emph{Advances in Neural Information Processing Systems}, volume~31. Curran Associates, Inc., 2018.
\newblock URL \url{https://proceedings.neurips.cc/paper_files/paper/2018/file/139c3c1b7ca46a9d4fd6d163d98af635-Paper.pdf}.

\bibitem[De~Lange et~al.(2022)De~Lange, Aljundi, Masana, Parisot, Jia, Leonardis, Slabaugh, and Tuytelaars]{delange2022continuallearning}
Matthias De~Lange, Rahaf Aljundi, Marc Masana, Sarah Parisot, Xu~Jia, Ales Leonardis, Gregory Slabaugh, and Tinne Tuytelaars.
\newblock A continual learning survey: Defying forgetting in classification tasks.
\newblock \emph{IEEE transactions on pattern analysis and machine intelligence}, 44\penalty0 (7):\penalty0 3366--3385, July 2022.

\bibitem[Deng \& Papadimitriou(1994)Deng and Papadimitriou]{deng_papadimitriou1994complexitycoop}
Xiaotie Deng and Christos~H. Papadimitriou.
\newblock On the complexity of cooperative solution concepts.
\newblock 19:\penalty0 257–266, 1994.
\newblock ISSN 0364-765X.

\bibitem[Du et~al.(2022)Du, Abbeel, and Grover]{du2022curriculumselfplay}
Yuqing Du, P~Abbeel, and Aditya Grover.
\newblock It takes four to tango: Multiagent selfplay for automatic curriculum generation.
\newblock \emph{International Conference on Learning Representations}, 2022.

\bibitem[Dughmi(2009)]{dughmi2009submodularity}
Shaddin Dughmi.
\newblock Submodular functions: Extensions, distributions, and algorithms. a survey.
\newblock December 2009.

\bibitem[Elkind et~al.(2009)Elkind, Goldberg, Goldberg, and Wooldridge]{elkind2009solutionconceptscomputational}
Edith Elkind, Leslie~Ann Goldberg, Paul~W Goldberg, and Michael Wooldridge.
\newblock On the computational complexity of weighted voting games.
\newblock \emph{Annals of mathematics and artificial intelligence}, 56\penalty0 (2):\penalty0 109--131, June 2009.

\bibitem[Elman(1993)]{elman1993curriculum}
J~L Elman.
\newblock Learning and development in neural networks: the importance of starting small.
\newblock \emph{Cognition}, 48\penalty0 (1):\penalty0 71--99, July 1993.

\bibitem[Faigle(2022)]{faigle2022gametheory}
Ulrich Faigle.
\newblock \emph{Mathematical Game Theory}.
\newblock 2022.

\bibitem[Farnia \& Ozdaglar(2020)Farnia and Ozdaglar]{farnia2020gangames}
Farzan Farnia and Asuman Ozdaglar.
\newblock {GANs} may have no nash equilibria.
\newblock \emph{arXiv e-prints}, pp.\  arXiv:2002.09124, February 2020.

\bibitem[Fedus et~al.(2020)Fedus, Ramachandran, Agarwal, Bengio, Larochelle, Rowland, and Dabney]{fedus2020experiencereplay}
William Fedus, Prajit Ramachandran, Rishabh Agarwal, Yoshua Bengio, Hugo Larochelle, Mark Rowland, and Will Dabney.
\newblock Revisiting fundamentals of experience replay.
\newblock In Hal~Daumé III and Aarti Singh (eds.), \emph{Proceedings of the 37th International Conference on Machine Learning}, volume 119 of \emph{Proceedings of Machine Learning Research}, pp.\  3061--3071. PMLR, 13--18 Jul 2020.
\newblock URL \url{https://proceedings.mlr.press/v119/fedus20a.html}.

\bibitem[Feng et~al.(2021)Feng, Gomes, and Selman]{feng2021curriculumbandit}
Dieqiao Feng, Carla~P Gomes, and Bart Selman.
\newblock A novel automated curriculum strategy to solve hard sokoban planning instances.
\newblock October 2021.

\bibitem[Fifty et~al.(2021)Fifty, Amid, Zhao, Yu, Anil, and Finn]{Fifty2021-wi}
Christopher Fifty, E~Amid, Zhe Zhao, Tianhe Yu, Rohan Anil, and Chelsea Finn.
\newblock Efficiently identifying task groupings for {Multi-Task} learning.
\newblock \emph{Advances in neural information processing systems}, 2021.

\bibitem[Flood(1952)]{merrillflood1952pd}
Merrill~M. Flood.
\newblock \emph{Some Experimental Games}.
\newblock RAND Corporation, Santa Monica, CA, 1952.

\bibitem[Florensa et~al.(2017)Florensa, Held, Wulfmeier, Zhang, and Abbeel]{florensa2017reversecl}
Carlos Florensa, David Held, Markus Wulfmeier, Michael Zhang, and Pieter Abbeel.
\newblock Reverse curriculum generation for reinforcement learning.
\newblock In Sergey Levine, Vincent Vanhoucke, and Ken Goldberg (eds.), \emph{Proceedings of the 1st Annual Conference on Robot Learning}, volume~78 of \emph{Proceedings of Machine Learning Research}, pp.\  482--495. PMLR, 2017.

\bibitem[Florensa et~al.(2018)Florensa, Held, Geng, and Abbeel]{florensa2018curriculumgan}
Carlos Florensa, David Held, Xinyang Geng, and Pieter Abbeel.
\newblock Automatic goal generation for reinforcement learning agents.
\newblock In Jennifer Dy and Andreas Krause (eds.), \emph{Proceedings of the 35th International Conference on Machine Learning}, volume~80 of \emph{Proceedings of Machine Learning Research}, pp.\  1515--1528. PMLR, 2018.

\bibitem[F.R.S.(1901)]{pearson1901pca}
Karl~Pearson F.R.S.
\newblock Liii. on lines and planes of closest fit to systems of points in space.
\newblock \emph{The London, Edinburgh, and Dublin Philosophical Magazine and Journal of Science}, 2\penalty0 (11):\penalty0 559--572, 1901.
\newblock \doi{10.1080/14786440109462720}.
\newblock URL \url{https://doi.org/10.1080/14786440109462720}.

\bibitem[Garnelo et~al.(2021)Garnelo, Czarnecki, Liu, Tirumala, Oh, Gidel, van Hasselt, and Balduzzi]{garnelo2021pick}
Marta Garnelo, Wojciech~Marian Czarnecki, Siqi Liu, Dhruva Tirumala, Junhyuk Oh, Gauthier Gidel, Hado van Hasselt, and David Balduzzi.
\newblock Pick your battles: Interaction graphs as population-level objectives for strategic diversity.
\newblock In \emph{Proceedings of the 20th International Conference on Autonomous Agents and MultiAgent Systems}, AAMAS '21, pp.\  1501–1503, Richland, SC, 2021. International Foundation for Autonomous Agents and Multiagent Systems.
\newblock ISBN 9781450383073.

\bibitem[Gemp et~al.(2021)Gemp, McWilliams, Vernade, and Graepel]{gemp2021eigengame}
Ian Gemp, Brian McWilliams, Claire Vernade, and Thore Graepel.
\newblock Eigengame: {\{}PCA{\}} as a nash equilibrium.
\newblock In \emph{International Conference on Learning Representations}, 2021.
\newblock URL \url{https://openreview.net/forum?id=NzTU59SYbNq}.

\bibitem[Ghorbani \& Zou(2019)Ghorbani and Zou]{ghorbani2019datashapley}
Amirata Ghorbani and James Zou.
\newblock Data shapley: Equitable valuation of data for machine learning.
\newblock April 2019.

\bibitem[Gittins(1979)]{gittins1979bandits}
J~C Gittins.
\newblock Bandit processes and dynamic allocation indices.
\newblock \emph{Journal of the Royal Statistical Society}, 41\penalty0 (2):\penalty0 148--164, January 1979.

\bibitem[Goodfellow et~al.(2014)Goodfellow, Pouget-Abadie, Mirza, Xu, Warde-Farley, Ozair, Courville, and Bengio]{goodfellow2014gans}
Ian Goodfellow, Jean Pouget-Abadie, Mehdi Mirza, Bing Xu, David Warde-Farley, Sherjil Ozair, Aaron Courville, and Yoshua Bengio.
\newblock Generative adversarial nets.
\newblock In Z.~Ghahramani, M.~Welling, C.~Cortes, N.~Lawrence, and K.Q. Weinberger (eds.), \emph{Advances in Neural Information Processing Systems}, volume~27. Curran Associates, Inc., 2014.
\newblock URL \url{https://proceedings.neurips.cc/paper_files/paper/2014/file/5ca3e9b122f61f8f06494c97b1afccf3-Paper.pdf}.

\bibitem[Goodfellow et~al.(2016)Goodfellow, Bengio, and Courville]{goodfellow2016deeplearningbook}
Ian Goodfellow, Yoshua Bengio, and Aaron Courville.
\newblock \emph{Deep learning}.
\newblock MIT press, 2016.

\bibitem[Grabisch \& Roubens(1999)Grabisch and Roubens]{Grabisch_Roubens_1999}
Michel Grabisch and Marc Roubens.
\newblock An axiomatic approach to the concept of interaction among players in cooperative games.
\newblock \emph{International Journal of Game Theory}, 28\penalty0 (4):\penalty0 547–565, Nov 1999.
\newblock ISSN 0020-7276.

\bibitem[Graves et~al.(2017)Graves, Bellemare, Menick, Munos, and Kavukcuoglu]{graves2017curriculumbandit}
Alex Graves, Marc~G Bellemare, Jacob Menick, R{\'e}mi Munos, and Koray Kavukcuoglu.
\newblock Automated curriculum learning for neural networks.
\newblock In Doina Precup and Yee~Whye Teh (eds.), \emph{Proceedings of the 34th International Conference on Machine Learning}, volume~70 of \emph{Proceedings of Machine Learning Research}, pp.\  1311--1320. PMLR, 2017.

\bibitem[Hausken \& Mohr(2001)Hausken and Mohr]{hausken_mohr_2001shapleyothers}
Kjell Hausken and Matthias Mohr.
\newblock The value of a player in n-person games.
\newblock \emph{Social choice and welfare}, 18\penalty0 (3):\penalty0 465–483, 2001.
\newblock ISSN 0176-1714.

\bibitem[Hilbe et~al.(2013)Hilbe, Nowak, and Sigmund]{hilbe_nowak_sigmund_2013zerodet}
Christian Hilbe, Martin~A. Nowak, and Karl Sigmund.
\newblock Evolution of extortion in iterated prisoner’s dilemma games.
\newblock \emph{Proceedings of the National Academy of Sciences of the United States of America}, 110\penalty0 (17):\penalty0 6913–6918, Apr 2013.
\newblock ISSN 0027-8424.

\bibitem[Hotelling(1933)]{hotelling1933pca}
Harold Hotelling.
\newblock Analysis of a complex of statistical variables into principal components.
\newblock \emph{Journal of Educational Psychology}, 24:\penalty0 498--520, 1933.

\bibitem[Huang et~al.(2022{\natexlab{a}})Huang, Xu, Fang, and Zhao]{huang2022adverarialrl}
Peide Huang, Mengdi Xu, Fei Fang, and Ding Zhao.
\newblock Robust reinforcement learning as a stackelberg game via adaptively-regularized adversarial training.
\newblock February 2022{\natexlab{a}}.

\bibitem[Huang et~al.(2022{\natexlab{b}})Huang, Dossa, Ye, Braga, Chakraborty, Mehta, and Araújo]{huang2022cleanrl}
Shengyi Huang, Rousslan Fernand~Julien Dossa, Chang Ye, Jeff Braga, Dipam Chakraborty, Kinal Mehta, and João~G.M. Araújo.
\newblock Cleanrl: High-quality single-file implementations of deep reinforcement learning algorithms.
\newblock \emph{Journal of Machine Learning Research}, 23\penalty0 (274):\penalty0 1--18, 2022{\natexlab{b}}.
\newblock URL \url{http://jmlr.org/papers/v23/21-1342.html}.

\bibitem[Jin et~al.(2019)Jin, Netrapalli, and Jordan]{jin2019gameoptim}
Chi Jin, Praneeth Netrapalli, and Michael~I Jordan.
\newblock Minmax optimization: Stable limit points of gradient descent ascent are locally optimal.
\newblock \emph{arXiv.org}, 2019.

\bibitem[Karpathy \& Abbeel(2021)Karpathy and Abbeel]{Karpathy2021tesla}
Andrej Karpathy and Pieter Abbeel.
\newblock The robot brains podcast: Andrej karpathy on the visionary ai in tesla's autonomous driving.
\newblock \url{https://podcasts.apple.com/us/podcast/andrej-karpathy-on-visionary-ai-in-teslas-autonomous/id1559275284?i=1000513993723}, 2021.

\bibitem[Kingma \& Ba(2015)Kingma and Ba]{kignma_ba14adam}
Diederik~P. Kingma and Jimmy Ba.
\newblock Adam: {A} method for stochastic optimization.
\newblock In Yoshua Bengio and Yann LeCun (eds.), \emph{3rd International Conference on Learning Representations, {ICLR} 2015, San Diego, CA, USA, May 7-9, 2015, Conference Track Proceedings}, 2015.
\newblock URL \url{http://arxiv.org/abs/1412.6980}.

\bibitem[Knight et~al.(2021)Knight, Campbell, {Marc}, Gaffney, Shaw, Janga, Glynatsi, Campbell, Langner, Singh, Rymer, Campbell, Young, {MHakem}, Palmer, Glass, Mancia, {edouardArgenson}, Jones, {kjurgielajtis}, Murase, Parvatikar, Beck, Davidson-Pilon, Zoulias, Pohl, Slavin, Standen, Kratz, and Ahmed]{knight2021pyaxelrod}
Vince Knight, Owen Campbell, {Marc}, T~J Gaffney, Eric Shaw, Vsn~Reddy Janga, Nikoleta Glynatsi, James Campbell, Karol~M Langner, Sourav Singh, Julie Rymer, Thomas Campbell, Jason Young, {MHakem}, Geraint Palmer, Kristian Glass, Daniel Mancia, {edouardArgenson}, Martin Jones, {kjurgielajtis}, Yohsuke Murase, Sudarshan Parvatikar, Melanie Beck, Cameron Davidson-Pilon, Marios Zoulias, Adam Pohl, Paul Slavin, Timothy Standen, Aaron Kratz, and Areeb Ahmed.
\newblock {Axelrod-Python/Axelrod}: v4.12.0, October 2021.

\bibitem[Krause \& Guestrin(2011)Krause and Guestrin]{krause2011submodularity}
Andreas Krause and Carlos Guestrin.
\newblock Submodularity and its applications in optimized information gathering.
\newblock \emph{ACM Trans. Intell. Syst. Technol.}, 2\penalty0 (4):\penalty0 1--20, July 2011.

\bibitem[Krizhevsky(2009)]{krizhevsky2009cifardatesets}
Alex Krizhevsky.
\newblock Learning multiple layers of features from tiny images.
\newblock Technical report, April 2009.

\bibitem[Krizhevsky et~al.(2009)Krizhevsky, Hinton, et~al.]{krizhevsky2009learning}
Alex Krizhevsky, Geoffrey Hinton, et~al.
\newblock Learning multiple layers of features from tiny images.
\newblock 2009.

\bibitem[Krueger \& Dayan(2009)Krueger and Dayan]{krueger_dayan2009curriculum}
Kai~A Krueger and Peter Dayan.
\newblock Flexible shaping: how learning in small steps helps.
\newblock \emph{Cognition}, 110\penalty0 (3):\penalty0 380--394, March 2009.

\bibitem[Lanctot et~al.(2017)Lanctot, Zambaldi, Gruslys, Lazaridou, Tuyls, P{\'e}rolat, Silver, and Graepel]{lanctot2017unified}
Marc Lanctot, Vinicius Zambaldi, Audrunas Gruslys, Angeliki Lazaridou, Karl Tuyls, Julien P{\'e}rolat, David Silver, and Thore Graepel.
\newblock A unified game-theoretic approach to multiagent reinforcement learning.
\newblock \emph{Advances in neural information processing systems}, 30, 2017.

\bibitem[Lattimore \& Szepesv{\'a}ri(2020)Lattimore and Szepesv{\'a}ri]{lattimore2020bandits}
Tor Lattimore and Csaba Szepesv{\'a}ri.
\newblock \emph{Bandit Algorithms}.
\newblock Cambridge University Press, July 2020.

\bibitem[LeCun \& Cortes(2010)LeCun and Cortes]{lecun-mnisthandwrittendigit-2010}
Yann LeCun and Corinna Cortes.
\newblock {MNIST} handwritten digit database.
\newblock 2010.
\newblock URL \url{http://yann.lecun.com/exdb/mnist/}.

\bibitem[Lee et~al.(2021)Lee, Goldt, and Saxe]{Lee_Goldt_Saxe_2021}
Sebastian Lee, Sebastian Goldt, and Andrew~M. Saxe.
\newblock Continual learning in the teacher-student setup: Impact of task similarity.
\newblock \emph{International Conference on Machine Learning}, 2021.
\newblock URL \url{https://www.semanticscholar.org/paper/57db7f24f15150ef7ea0db1fed20e6ee752792ec}.

\bibitem[Lin(1992)]{lin1992experiencereplay}
Long-Ji Lin.
\newblock Self-improving reactive agents based on reinforcement learning, planning and teaching.
\newblock \emph{Machine learning}, 8\penalty0 (3):\penalty0 293--321, May 1992.

\bibitem[Lin et~al.(2019)Lin, Zhen, Li, Zhang, and Kwong]{Lin2019-ew}
Xi~Lin, Hui-Ling Zhen, Zhenhua Li, Qingfu Zhang, and S~Kwong.
\newblock Pareto {Multi-Task} learning.
\newblock \emph{Advances in neural information processing systems}, 2019.

\bibitem[Lin et~al.(2021)Lin, Shi, Pathak, and Ramanan]{lin2021clear}
Zhiqiu Lin, Jia Shi, Deepak Pathak, and Deva Ramanan.
\newblock The clear benchmark: Continual learning on real-world imagery.
\newblock In \emph{Thirty-fifth Conference on Neural Information Processing Systems Datasets and Benchmarks Track (Round 2)}, 2021.

\bibitem[Liu et~al.(2020)Liu, Wu, and Mozafari]{liu2020curriculumbanditadam}
Rui Liu, Tianyi Wu, and Barzan Mozafari.
\newblock Adam with bandit sampling for deep learning.
\newblock October 2020.

\bibitem[Liu et~al.(2022)Liu, Marris, Hennes, Merel, Heess, and Graepel]{liu2022neupl}
Siqi Liu, Luke Marris, Daniel Hennes, Josh Merel, Nicolas Heess, and Thore Graepel.
\newblock Neu{PL}: Neural population learning.
\newblock In \emph{International Conference on Learning Representations}, 2022.
\newblock URL \url{https://openreview.net/forum?id=MIX3fJkl_1}.

\bibitem[Lomonaco \& Maltoni(2017)Lomonaco and Maltoni]{lomonaco17core50}
Vincenzo Lomonaco and Davide Maltoni.
\newblock Core50: a new dataset and benchmark for continuous object recognition.
\newblock In Sergey Levine, Vincent Vanhoucke, and Ken Goldberg (eds.), \emph{Proceedings of the 1st Annual Conference on Robot Learning}, volume~78 of \emph{Proceedings of Machine Learning Research}, pp.\  17--26. PMLR, 13--15 Nov 2017.
\newblock URL \url{https://proceedings.mlr.press/v78/lomonaco17a.html}.

\bibitem[Lundberg \& Lee(2017)Lundberg and Lee]{lundberg2017shap}
Scott~M. Lundberg and Su-In Lee.
\newblock A unified approach to interpreting model predictions.
\newblock In \emph{Proceedings of the 31st International Conference on Neural Information Processing Systems}, NIPS'17, pp.\  4768–4777, Red Hook, NY, USA, 2017. Curran Associates Inc.
\newblock ISBN 9781510860964.

\bibitem[Madry et~al.(2018)Madry, Makelov, Schmidt, Tsipras, and Vladu]{madry2018adversarialattacks}
Aleksander Madry, Aleksandar Makelov, Ludwig Schmidt, Dimitris Tsipras, and Adrian Vladu.
\newblock Towards deep learning models resistant to adversarial attacks.
\newblock In \emph{International Conference on Learning Representations}, 2018.
\newblock URL \url{https://openreview.net/forum?id=rJzIBfZAb}.

\bibitem[Matiisen et~al.(2020)Matiisen, Oliver, Cohen, and Schulman]{matiisen2020curriculumbandit}
Tambet Matiisen, Avital Oliver, Taco Cohen, and John Schulman.
\newblock {Teacher-Student} curriculum learning.
\newblock \emph{IEEE transactions on neural networks and learning systems}, 31\penalty0 (9):\penalty0 3732--3740, September 2020.

\bibitem[McCloskey \& Cohen(1989)McCloskey and Cohen]{mccloskey1989catastrophic}
Michael McCloskey and Neal~J Cohen.
\newblock Catastrophic interference in connectionist networks: The sequential learning problem.
\newblock In Gordon~H Bower (ed.), \emph{Psychology of Learning and Motivation}, volume~24, pp.\  109--165. Academic Press, January 1989.

\bibitem[Mitchell et~al.(2022)Mitchell, Cooper, Frank, and Holmes]{mitchell2022samplingshapley}
Rory Mitchell, Joshua Cooper, Eibe Frank, and Geoffrey Holmes.
\newblock Sampling permutations for shapley value estimation.
\newblock \emph{Journal of machine learning research: JMLR}, 23\penalty0 (43):\penalty0 1--46, 2022.

\bibitem[Mohebbi~Moghaddam et~al.(2023)Mohebbi~Moghaddam, Boroomand, Jalali, Zareian, Daeijavad, Manshaei, and Krunz]{mohebbi2023gangames}
Monireh Mohebbi~Moghaddam, Bahar Boroomand, Mohammad Jalali, Arman Zareian, Alireza Daeijavad, Mohammad~Hossein Manshaei, and Marwan Krunz.
\newblock Games of {GANs}: game-theoretical models for generative adversarial networks.
\newblock \emph{Artificial Intelligence Review}, February 2023.

\bibitem[Mundt et~al.(2023)Mundt, Hong, Pliushch, and Ramesh]{mundt2023continuallearning}
Martin Mundt, Yongwon Hong, Iuliia Pliushch, and Visvanathan Ramesh.
\newblock A wholistic view of continual learning with deep neural networks: Forgotten lessons and the bridge to active and open world learning.
\newblock \emph{Neural networks: the official journal of the International Neural Network Society}, 160:\penalty0 306--336, March 2023.

\bibitem[Narvekar et~al.(2022)Narvekar, Peng, Leonetti, Sinapov, Taylor, and Stone]{narvekar2022surveyclrl}
Sanmit Narvekar, Bei Peng, Matteo Leonetti, Jivko Sinapov, Matthew~E Taylor, and Peter Stone.
\newblock Curriculum learning for reinforcement learning domains: a framework and survey.
\newblock \emph{Journal of machine learning research: JMLR}, 21\penalty0 (1):\penalty0 7382--7431, June 2022.

\bibitem[Nash(1950)]{nash1950equilibrium}
J~F Nash.
\newblock Equilibrium points in {N-Person} games.
\newblock \emph{Proceedings of the National Academy of Sciences of the United States of America}, 36\penalty0 (1):\penalty0 48--49, January 1950.

\bibitem[Ng(2021)]{ng2021data}
Andrew Ng.
\newblock Mlops: From model-centric to data-centric ai.
\newblock \url{https://www.youtube.com/watch?v=06-AZXmwHjo}, 2021.

\bibitem[Nowak \& Radzik(1994)Nowak and Radzik]{Nowak1994shapleygen}
Andrzej~S Nowak and Tadeusz Radzik.
\newblock The shapley value for n-person games in generalized characteristic function form.
\newblock \emph{Games and economic behavior}, 6\penalty0 (1):\penalty0 150--161, January 1994.

\bibitem[Oudeyer et~al.(2007)Oudeyer, Kaplan, and Hafner]{oudeyer2007intrinsic}
Pierre-Yves Oudeyer, Fr\'ed\'eric Kaplan, and Verena~V Hafner.
\newblock Intrinsic motivation systems for autonomous mental development.
\newblock \emph{IEEE transactions on evolutionary computation}, 11\penalty0 (2):\penalty0 265--286, 2007.

\bibitem[Parisi et~al.(2019)Parisi, Kemker, Part, Kanan, and Wermter]{Parisi2019continuallearning}
German~I Parisi, Ronald Kemker, Jose~L Part, Christopher Kanan, and Stefan Wermter.
\newblock Continual lifelong learning with neural networks: A review.
\newblock \emph{Neural networks: the official journal of the International Neural Network Society}, 113:\penalty0 54--71, May 2019.

\bibitem[Paszke et~al.(2019)Paszke, Gross, Massa, Lerer, Bradbury, Chanan, Killeen, Lin, Gimelshein, Antiga, Desmaison, Kopf, Yang, DeVito, Raison, Tejani, Chilamkurthy, Steiner, Fang, Bai, and Chintala]{paszke2019pytorch}
Adam Paszke, Sam Gross, Francisco Massa, Adam Lerer, James Bradbury, Gregory Chanan, Trevor Killeen, Zeming Lin, Natalia Gimelshein, Luca Antiga, Alban Desmaison, Andreas Kopf, Edward Yang, Zachary DeVito, Martin Raison, Alykhan Tejani, Sasank Chilamkurthy, Benoit Steiner, Lu~Fang, Junjie Bai, and Soumith Chintala.
\newblock Pytorch: An imperative style, high-performance deep learning library.
\newblock In \emph{Advances in Neural Information Processing Systems 32}, pp.\  8024--8035. Curran Associates, Inc., 2019.
\newblock URL \url{http://papers.neurips.cc/paper/9015-pytorch-an-imperative-style-high-performance-deep-learning-library.pdf}.

\bibitem[Patel et~al.(2021)Patel, Garnelo, Gemp, Dyer, and Bachrach]{romapatel2021deepmind}
Roma Patel, Marta Garnelo, Ian Gemp, Chris Dyer, and Yoram Bachrach.
\newblock Game-theoretic vocabulary selection via the shapley value and banzhaf index.
\newblock In \emph{Proceedings of the 2021 Conference of the North American Chapter of the Association for Computational Linguistics: Human Language Technologies}, pp.\  2789--2798, Online, June 2021. Association for Computational Linguistics.

\bibitem[Portelas et~al.(2019{\natexlab{a}})Portelas, Colas, Hofmann, and Oudeyer]{portelas2019curriculumbandit}
R{\'e}my Portelas, C{\'e}dric Colas, Katja Hofmann, and Pierre-Yves Oudeyer.
\newblock Teacher algorithms for curriculum learning of deep {RL} in continuously parameterized environments.
\newblock October 2019{\natexlab{a}}.

\bibitem[Portelas et~al.(2019{\natexlab{b}})Portelas, Colas, Hofmann, and Oudeyer]{portelas2019teachercurriculum}
R{\'e}my Portelas, C{\'e}dric Colas, Katja Hofmann, and Pierre-Yves Oudeyer.
\newblock Teacher algorithms for curriculum learning of deep {RL} in continuously parameterized environments.
\newblock October 2019{\natexlab{b}}.

\bibitem[Portelas et~al.(2020)Portelas, Romac, Hofmann, and Oudeyer]{portelas2020metacurriculum}
R{\'e}my Portelas, Cl{\'e}ment Romac, Katja Hofmann, and Pierre-Yves Oudeyer.
\newblock Meta automatic curriculum learning.
\newblock November 2020.

\bibitem[Procaccia et~al.(2014)Procaccia, Shah, and Tucker]{Procaccia_Shah_Tucker_2014}
Ariel Procaccia, Nisarg Shah, and Max Tucker.
\newblock On the structure of synergies in cooperative games.
\newblock \emph{Proceedings of the AAAI Conference on Artificial Intelligence}, 28\penalty0 (1), Jun 2014.
\newblock ISSN 2374-3468.
\newblock \doi{10.1609/aaai.v28i1.8812}.
\newblock URL \url{https://ojs.aaai.org/index.php/AAAI/article/view/8812}.

\bibitem[Racaniere et~al.(2020)Racaniere, Lampinen, Santoro, Reichert, Firoiu, and Lillicrap]{Racaniere2020curriculumsettersolver}
Sebastien Racaniere, Andrew Lampinen, Adam Santoro, David Reichert, Vlad Firoiu, and Timothy Lillicrap.
\newblock Automated curriculum generation through setter-solver interactions.
\newblock In \emph{International Conference on Learning Representations}, 2020.

\bibitem[Romac et~al.(2021)Romac, Portelas, Hofmann, and Oudeyer]{romac2021curriculum}
Cl{\'e}ment Romac, R{\'e}my Portelas, Katja Hofmann, and Pierre-Yves Oudeyer.
\newblock {TeachMyAgent}: a benchmark for automatic curriculum learning in deep {RL}.
\newblock March 2021.

\bibitem[Roth(1988)]{roth1988shapley}
Alvin~E Roth (ed.).
\newblock \emph{The Shapley value: essays in honor of Lloyd S. Shapley}.
\newblock Cambridge University Press, October 1988.

\bibitem[Sanchez \& Berganti{\~n}os(1997)Sanchez and Berganti{\~n}os]{Sanchez1997shapleygen}
Estela Sanchez and Gustavo Berganti{\~n}os.
\newblock On values for generalized characteristic functions.
\newblock \emph{Operations-Research-Spektrum}, 19\penalty0 (3):\penalty0 229--234, September 1997.

\bibitem[Schaul et~al.(2016)Schaul, Quan, Antonoglou, and Silver]{schaul2016per}
Tom Schaul, John Quan, Ioannis Antonoglou, and David Silver.
\newblock Prioritized experience replay.
\newblock In \emph{4th International Conference on Learning Representations}, 2016.

\bibitem[Schmidhuber(1991)]{Schmidhuber_1991}
J.~Schmidhuber.
\newblock \emph{A possibility for implementing curiosity and boredom in model-building neural controllers}.
\newblock The MIT Press, 1991.
\newblock ISBN 9780262256674.

\bibitem[Schulman et~al.(2017)Schulman, Wolski, Dhariwal, Radford, and Klimov]{schulman2017proximal}
John Schulman, Filip Wolski, Prafulla Dhariwal, Alec Radford, and Oleg Klimov.
\newblock Proximal policy optimization algorithms.
\newblock \emph{arXiv preprint arXiv:1707.06347}, 2017.

\bibitem[Settles(2010)]{settles2010activelearning}
Burr Settles.
\newblock Active learning literature survey.
\newblock \emph{Machine learning}, 15\penalty0 (2):\penalty0 201--221, 2010.

\bibitem[Shamsian et~al.(2023)Shamsian, Navon, Glazer, Kawaguchi, Chechik, and Fetaya]{Shamsian2023-gf}
Aviv Shamsian, Aviv Navon, Neta Glazer, Kenji Kawaguchi, Gal Chechik, and Ethan Fetaya.
\newblock Auxiliary learning as an asymmetric bargaining game.
\newblock \emph{ArXiv}, 2023.

\bibitem[Shapley(1952)]{shapley1952value}
Lloyd~S Shapley.
\newblock A value for {N-Person} games.
\newblock Technical report, RAND Corporation, 1952.

\bibitem[Shapley(1971)]{shapley1971coreconvex}
Lloyd~S Shapley.
\newblock Cores of convex games.
\newblock \emph{International Journal of Game Theory}, 1\penalty0 (1):\penalty0 11--26, December 1971.

\bibitem[Shubik(1981)]{shubik1981solutions}
Martin Shubik.
\newblock Game theory models and methods in political economy.
\newblock In Kenneth~J Arrow and Michael~D Intriligator (eds.), \emph{Handbook of Mathematical Economics}, volume~1, pp.\  285--330. Elsevier, 1981.

\bibitem[Soviany et~al.(2022)Soviany, Ionescu, Rota, and Sebe]{soviany2022surveycl}
Petru Soviany, Radu~Tudor Ionescu, Paolo Rota, and Nicu Sebe.
\newblock Curriculum learning: A survey.
\newblock \emph{International journal of computer vision}, 130\penalty0 (6):\penalty0 1526--1565, June 2022.

\bibitem[Srinivasan et~al.(2022)Srinivasan, Chang, Pinto~Alva, Chochlakis, Rostami, and Thomason]{srinivasan2022climb}
Tejas Srinivasan, Ting-Yun Chang, Leticia Pinto~Alva, Georgios Chochlakis, Mohammad Rostami, and Jesse Thomason.
\newblock Climb: A continual learning benchmark for vision-and-language tasks.
\newblock volume~35, pp.\  29440--29453, 2022.

\bibitem[Standley et~al.(2019)Standley, Zamir, Chen, Guibas, Malik, and Savarese]{Standley2019-sk}
Trevor Standley, Amir~R Zamir, Dawn Chen, Leonidas Guibas, Jitendra Malik, and Silvio Savarese.
\newblock Which tasks should be learned together in multi-task learning?
\newblock May 2019.

\bibitem[Sutton \& Barto(2018)Sutton and Barto]{sutton2018rlintro}
Richard~S Sutton and Andrew~G Barto.
\newblock \emph{Reinforcement Learning: An Introduction}.
\newblock The MIT Press, 2nd edition, 2018.

\bibitem[Turchetta et~al.(2020)Turchetta, Kolobov, Shah, Krause, and Agarwal]{turchetta2020curriculumbandirsaferl}
Matteo Turchetta, Andrey Kolobov, Shital Shah, Andreas Krause, and Alekh Agarwal.
\newblock Safe reinforcement learning via curriculum induction.
\newblock \emph{Advances in neural information processing systems}, 33:\penalty0 12151--12162, 2020.

\bibitem[van~den Brink \& van~der Laan(1998)van~den Brink and van~der Laan]{vandebnrink1998shapleyaxioms}
René van~den Brink and Gerard van~der Laan.
\newblock Axiomatizations of the normalized banzhaf value and the shapley value.
\newblock \emph{Social Choice and Welfare}, 15\penalty0 (4):\penalty0 567--582, 1998.
\newblock ISSN 01761714, 1432217X.
\newblock URL \url{http://www.jstor.org/stable/41106281}.

\bibitem[Von~Neumann \& Morgenstern(1944)Von~Neumann and Morgenstern]{vonneumann1944theory}
John Von~Neumann and Oskar Morgenstern.
\newblock \emph{Theory of games and economic behavior}.
\newblock Princeton University Press, Princeton, NJ, 1944.

\bibitem[Wang et~al.(2022)Wang, Chen, and Zhu]{wang2022surveycl}
Xin Wang, Yudong Chen, and Wenwu Zhu.
\newblock A survey on curriculum learning.
\newblock \emph{IEEE transactions on pattern analysis and machine intelligence}, 44\penalty0 (9):\penalty0 4555--4576, September 2022.

\bibitem[Weber(1988)]{weber1988probshapley}
Robert~J Weber.
\newblock Probabilistic values for games.
\newblock In A~E Roth (ed.), \emph{Essays on the Shapley Value and Its Applications}, pp.\  101--119. Cambridge University Press, 1988.

\bibitem[Weinshall et~al.(2018)Weinshall, Cohen, and Amir]{weinshall2018curriculum}
Daphna Weinshall, Gad Cohen, and Dan Amir.
\newblock Curriculum learning by transfer learning: Theory and experiments with deep networks.
\newblock Technical report, 2018.

\bibitem[Wellman(2006)]{wellman2006egt}
Michael~P. Wellman.
\newblock Methods for empirical game-theoretic analysis.
\newblock In \emph{Proceedings of the 21st National Conference on Artificial Intelligence - Volume 2}, AAAI'06, pp.\  1552–1555. AAAI Press, 2006.
\newblock ISBN 9781577352815.

\bibitem[Wu et~al.(2020)Wu, Dyer, and Neyshabur]{Wu2020-hq}
Xiaoxia Wu, Ethan Dyer, and Behnam Neyshabur.
\newblock When do curricula work?
\newblock \emph{International Conference on Learning Representations}, 2020.

\bibitem[Yadan(2019)]{Yadan2019Hydra}
Omry Yadan.
\newblock Hydra - a framework for elegantly configuring complex applications.
\newblock Github, 2019.
\newblock URL \url{https://github.com/facebookresearch/hydra}.

\bibitem[Yan \& Procaccia(2021)Yan and Procaccia]{yanprocaccia2021shapleylovecore}
Tom Yan and Ariel~D Procaccia.
\newblock If you like shapley then you'll love the core.
\newblock \emph{Proceedings of the ... AAAI Conference on Artificial Intelligence. AAAI Conference on Artificial Intelligence}, 35\penalty0 (6):\penalty0 5751--5759, May 2021.

\bibitem[Zhang et~al.(2021)Zhang, Hayes, and Kanan]{Zhang2021-ip}
Yipeng Zhang, Tyler~L Hayes, and Christopher Kanan.
\newblock Disentangling transfer and interference in {Multi-Domain} learning.
\newblock \emph{ArXiv}, 2021.

\end{thebibliography}
\bibliographystyle{tmlr}

\newpage
\appendix

\section{Prospect Prior Experiments Details}
\label{sec:prospect-prior-hparams}
We provide for all problems, models or policies architectures, algorithm hyperparameters, and other reproducibility details.\footnotemark 
All models and architectures are implemented with \textsc{Pytorch}~\citep{paszke2019pytorch}, are configured using \textsc{Hydra}~\citep{Yadan2019Hydra}, and fit on a workstation equipped with a $16$ GB \textit{NVIDIA RTX A4000} GPU, $32$ GB of {RAM}, and $32$ CPU cores.

\footnotetext{Regardless, we plan to release the complete source code of all our experiments.}

\subsection{Supervised Learning}

\textbf{MNIST.}  We trained a model on the \textsc{Mnist}~\citep{lecun-mnisthandwrittendigit-2010} supervised \textit{10-digits} classification task. Specification of the model architecture and hyper-parameters selection are provided in ~\autoref{tab:mnist}.

\begin{table}[!htb]
    \centering
    \small
    \begin{subtable}[t]{0.49\linewidth}
        \begin{tabular}[t]{c|c}
            \toprule
             \textbf{Hyperparameter} & \textbf{Value}  \\ \midrule
             \textit{optimizer} & \textsc{Adam}~\citep{kignma_ba14adam} \\
             \textit{learning-rate} & $10^{-4}$\\
             \textit{betas} & $(0.9, 0.999)$ \\
             \textit{eps} & $10^{-8}$ \\
             \textit{batch-size} & $4$ \\
             \textit{epochs} & $200$ \\
             \textit{shuffle} & \textit{Yes} \\
             \bottomrule
        \end{tabular}
        \caption*{}
    \end{subtable}
    \begin{subtable}[t]{0.49\linewidth}
        \centering
        \begin{tabular}[t]{c}
            \toprule
             \textbf{Model} \\ \midrule
              ${\textsc{Conv2d} (32, 3, 1)}$ \\
              {\textsc{ReLU}()} \\
              {\textsc{Conv2d} (64, 3, 1)} \\
              {\textsc{ReLU}()} \\
              {\textsc{MaxPool2d} (2, 2)} \\
              {\textsc{DropOut}(0.25)} \\
             {\textsc{Flatten}()} \\
             \textsc{Linear}(9216, 128) \\
              {\textsc{ReLU}()} \\
              {\textsc{DropOut}(0.5)} \\
              \textsc{Linear}(128, 10) \\
             \bottomrule
        \end{tabular}
        \caption*{}
    \end{subtable}
    \caption{Details on the learning algorithm hyperparameters (\textit{left}) and model architecture (\textit{right}) used in the \textsc{MNIST}~\citep{lecun-mnisthandwrittendigit-2010} experiments. Model components and the optimizer are provided by \textsc{Pytorch}~\citep{paszke2019pytorch}. These details remained constant throughout the rest of the experiments with \textsc{MNIST}.} \label{tab:mnist}
\end{table}

\textbf{CIFAR10.} The experiments on \textsc{Cifar10}~\citep{krizhevsky2009learning} follow the same setting as those on \textsc{MNIST}. We similarly trained a model on the supervised \textit{10-classes} task. Specification of the model architecture and hyperparameter selection are provided in ~\autoref{tab:cifar10}. 

\begin{table}[!htb]
    \centering
    \small
    \begin{subtable}[t]{0.49\linewidth}
        \begin{tabular}[t]{c|c}
            \toprule
             \textbf{Hyperparameter} & \textbf{Value}  \\ \midrule
             \textit{optimizer} & \textsc{Adam}~\citep{kignma_ba14adam} \\
             \textit{learning-rate} & $10^{-4}$\\
             \textit{betas} & $(0.9, 0.999)$ \\
             \textit{eps} & $10^{-8}$ \\
             \textit{batch-size} & $4$ \\
             \textit{epochs} & $200$ \\
             \textit{shuffle} & \textit{Yes} \\
             \bottomrule
        \end{tabular}
        \caption*{\small}
    \end{subtable}
    \begin{subtable}[t]{0.49\linewidth}
        \centering
        \begin{tabular}[t]{c}
            \toprule
             \textbf{Model} \\ \midrule
              {\textsc{Conv2d} (3, 6, 5)} \\
             {\textsc{ReLU}()} \\
             {\textsc{MaxPool2d} (2, 2)} \\
             {\textsc{Conv2d} (6, 16, 5)} \\
             {\textsc{ReLU}()} \\
             {\textsc{MaxPool2d} (2, 2)} \\
             {\textsc{Flatten}()} \\
             {\textsc{Linear}(400, 120)} \\
             {\textsc{ReLU}()} \\
             {\textsc{Linear}(120, 84)} \\
             {\textsc{ReLU}()} \\
             {\textsc{Linear}(84, 10)} \\
             \bottomrule
        \end{tabular}
        \caption*{\small}
        \label{tab:arch-mnist}
    \end{subtable}
    \caption{Details on the learning algorithm hyperparameters (\textit{left}) and model architecture (\textit{right}) used in the \textsc{Cifar10}~\citep{krizhevsky2009learning} experiments. Model components and the optimizer are provided by \textsc{Pytorch}~\citep{paszke2019pytorch}. These details remained constant throughout the rest of the experiments with \textsc{Cifar10}.} \label{tab:cifar10}
\end{table}
\label{sec:confusion-matrix}

\begin{figure}
    \centering
    \small
    \begin{subfigure}{\textwidth}
        \centering
        \begin{subfigure}{0.25\textwidth}
            \centering
            \includegraphics[width=\textwidth]{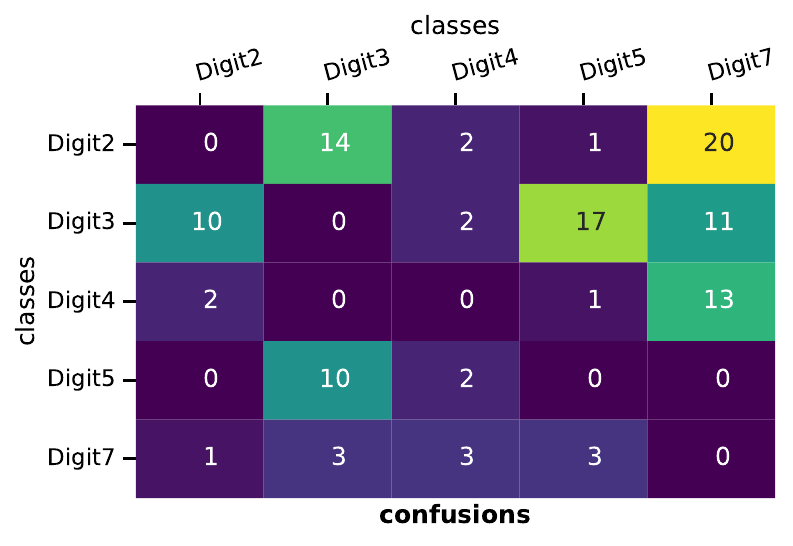}
        \end{subfigure}
        \begin{subfigure}{0.25\textwidth}
            \centering
            \includegraphics[width=\textwidth]{figures/experiments/shapley/MNIST.pdf}
        \end{subfigure}
        \begin{subfigure}{0.25\textwidth}
            \centering
            \includegraphics[width=\textwidth]{figures/experiments/vpop/MNIST.pdf}
        \end{subfigure}
        \caption{\tiny \textsc{MNIST}} \label{fig:mnist-truth}
    \end{subfigure}
    \begin{subfigure}{\textwidth}
        \centering
        \begin{subfigure}{0.25\textwidth}
            \centering
            \includegraphics[width=\textwidth]{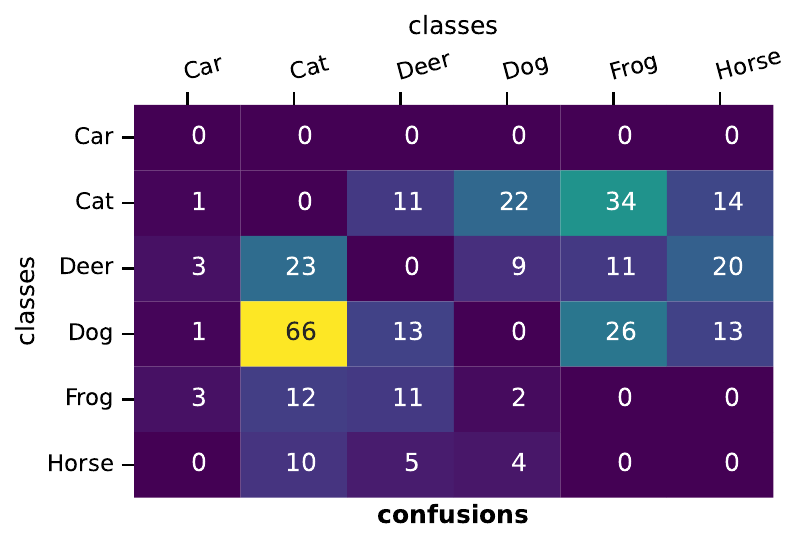}
        \end{subfigure}
        \begin{subfigure}{0.25\textwidth}
            \centering
            \includegraphics[width=\textwidth]{figures/experiments/shapley/CIFAR10.pdf}
        \end{subfigure}
        \begin{subfigure}{0.25\textwidth}
            \centering
            \includegraphics[width=\textwidth]{figures/experiments/vpop/CIFAR10.pdf}
        \end{subfigure}
        \caption{\tiny \textsc{CIFAR10}} \label{fig:cifar10-truth}
    \end{subfigure}

    \caption{\small The \textit{class-as-a-unit} analogy applied to \textsc{MNIST} (a) and \textsc{Cifar10} (b) served as our ground truth.
    For each problem, we derived the Shapley's value from the precomputed priors (\textit{left})~[\autoref{eq:shapley}] on each \textit{cooperative game} (\autoref{sec:expensive-prior}). Our results verify that units values on the \emph{target-unit} settings \textit{approximately} ordered the most confused pairs of classes. For instance, digits 2 \& 7 in \textsc{MNIST}, or \textit{dog} \& cat in \textsc{Cifar10}. When the target is \textit{all} classes, the \ac{vPoP} decomposition (\textit{right}) also (\autoref{sec:background-games}) identifies \textit{interfering} pairs.}
    \label{fig:my_label}
\end{figure}

\subsection{Reinforcement Learning}

\textsc{Minigrid Rooms.}  We utilized a sequence of  \textsc{TwoRooms}, \textsc{FourRooms}, and \textsc{SixRooms} \textit{gridworlds} provided on \textsc{Minigrid}~\citep{minigrid} as \textit{units of experience}. 
As the learning algorithm, we trained for $500,000$ steps a \textsc{PPO}~\cite{schulman2017proximal} agent, whose implementation we derived from \textsc{CleanRL}~\cite{huang2022cleanrl}. Policy and actor-critic architecture, with shared backbone, as well as other \textsc{PPO} hyperparameters details are presented in~\autoref{tab:minigridrooms}. For the \ac{TSCL} experiments, we leveraged \textit{Exp3S}~\citep{auer2003bandits} implementation from \citet{SMPyBandits} with default hyperparameters $\alpha=10^{-5}$ and $\gamma=0.05$, as defined in~\citet{graves2017curriculumbandit}.

\begin{table}[!htb]
    \centering
    \small
    \begin{subtable}[t]{0.49\linewidth}
        \centering
        \begin{tabular}[t]{c|c}
            \toprule
             \textbf{Hyperparameter} & \textbf{Value}  \\ \midrule
             \textit{optimizer} & \textsc{Adam}~\cite{kignma_ba14adam} \\
             \textit{learning-rate} & $0.0025$ \\
             \textit{annealing} & $\textit{Yes}$ \\
             \textit{num-steps} & $128$ \\
             \textit{total-timesteps} & $500,000$ \\
             \textit{seeds} & $5$ \\
             \textit{gamma} & $0.99$ \\
             \textit{GAE-lambda} & $0.95$ \\
             \textit{num-minibatches} & $4$ \\
             \textit{update-epochs} & $4$ \\
             \textit{advantage-normalization} & $\textit{Yes}$ \\ 
             \textit{clip-value-loss} & $\textit{Yes}$ \\
             \textit{clip-coeff} & $0.2$ \\
             \textit{entropy-coeff} & $0.01$ \\
             \textit{vf-coeff} & $0.5$ \\
             \textit{max-grad-norm} & $0.5$ \\
             \textit{target-kl} & \textit{No} \\
             \bottomrule
        \end{tabular}
        \caption*{}
    \end{subtable}
    \begin{subtable}[t]{0.49\linewidth}
        \centering
        \begin{tabular}[t]{c|c}
            \toprule
             \textbf{Actor} & \textbf{Critic}  \\ \midrule
             \multicolumn{2}{c}{\textsc{Conv2d}(16, 2, 2)} \\
             \multicolumn{2}{c}{\textsc{ReLU}()} \\
             \multicolumn{2}{c}{\textsc{MaxPool2d}(2, 2)} \\
             \multicolumn{2}{c}{\textsc{Conv2d}(16, 32, 2, 2)} \\
             \multicolumn{2}{c}{\textsc{ReLU}()} \\
             \multicolumn{2}{c}{\textsc{Conv2d}(16, 64, 2, 2)} \\
             \multicolumn{2}{c}{\textsc{ReLU}()} \\ \midrule
              $\textsc{Linear}(64, 64)$ & $\textsc{Linear}(64, 64)$ \\
              $\textsc{Tanh}()$ & $\textsc{Tanh}()$\\
              $\textsc{Linear}(64, 7)$ & $\textsc{Linear}(64, 1)$ \\
             \bottomrule
        \end{tabular}
        \caption*{}
    \end{subtable}
    \caption{Details on the \textsc{PPO} hyperparameters (\textit{left}) and \textit{actor-critic} architecture (\textit{right}) used in the \textsc{MinigridRooms}~\citep{minigrid} experiments. Policy and critic components, and the optimizer, are provided by \textsc{Pytorch}~\citep{paszke2019pytorch}. Implementation and default hyperparameters are derived from \textsc{CleanRL}~\citep{huang2022cleanrl}. These details remained constant throughout the rest of the experiments with \textsc{MinigridRooms}.}  \label{tab:minigridrooms}

\end{table}

\subsection{Populations \& Games.}

\textbf{Adversarial SIPD.} In our more challenging sparse and iterated version of Prisoner's Dilemma, at the end of $200$ interactions, inspired by Axelrod's competition~\citep{axelrod1981cooperation}. A \textit{win-draw-loss} reward $r = \{-1, 0, 1\}$ is given to a learning player if it beats a fixed opponent. Opponents are drawn from a population of five well-known strategies: \textit{always cooperate}, \textit{always defect}, \textit{win-stay-lose-switch}, \textit{tit-for-tat}, and a \textit{zero-determinant strategy}~\citep{axelrod1981cooperation, hilbe_nowak_sigmund_2013zerodet, knight2021pyaxelrod}. We trained for $500$ episodes (or $100,000$ steps) a \textsc{PPO}~\citep{schulman2017proximal} agent adapted from \textsc{CleanRL}~\cite{huang2022cleanrl} default implementation. Policy and actor-critic architecture, \textbf{without} shared backbone, as well as other \textsc{PPO} hyperparameters details are presented in~\autoref{tab:sipd}. For the \ac{TSCL} experiments, we leveraged \textit{Exp3S}~\citep{auer2003bandits} implementation from \citet{SMPyBandits} with default hyperparameters $\alpha=10^{-5}$ and $\gamma=0.05$, as defined in~\citet{graves2017curriculumbandit}.

\begin{table}[!htb]
    \centering
    \small
    \begin{subtable}[t]{0.49\linewidth}
        \centering
        \begin{tabular}[t]{c|c}
            \toprule
             \textbf{Hyperparameter} & \textbf{Value}  \\ \midrule
             \textit{optimizer} & \textsc{Adam}~\citep{kignma_ba14adam} \\
             \textit{learning-rate} & $0.0025$ \\
             \textit{annealing} & $\textit{Yes}$ \\
             \textit{num-steps} & $128$ \\
             \textit{timesteps} & $100,000$ \\
             \textit{seeds} & $5$ \\
             \textit{gamma} & $0.99$ \\
             \textit{GAE-lambda} & $0.95$ \\
             \textit{minibatches} & $4$ \\
             \textit{epochs} & $4$ \\
             \textit{advantage-norm} & $\textit{Yes}$ \\ 
             \textit{clip-value-loss} & $\textit{Yes}$ \\
             \textit{clip-coeff} & $0.2$ \\
             \textit{entropy-coeff} & $0.01$ \\
             \textit{vf-coeff} & $0.5$ \\
             \textit{max-grad-norm} & $0.5$ \\
             \textit{target-kl} & \textit{No} \\
             \bottomrule
        \end{tabular} \caption*{}
    \end{subtable}
    \begin{subtable}[t]{0.49\linewidth}
        \centering
        \begin{tabular}[t]{c|c}
            \toprule
             \textbf{Actor} & \textbf{Critic}  \\ \midrule
              $\textsc{Linear}(2, 64)$ & $\textsc{Linear}(2, 64)$\\
              $\textsc{OrthoInit}()$ & $\textsc{OrthoInit}()$\\
              $\textsc{Tanh}()$ &  $\textsc{Tanh}()$\\
              $\textsc{Linear}(64, 64)$ & $\textsc{Linear}(64, 64)$ \\
              $\textsc{OrthoInit}()$ & $\textsc{OrthoInit}()$\\
              $\textsc{Tanh}()$ & $\textsc{Tanh}()$\\
              $\textsc{Linear}(64, 2)$ & $\textsc{Linear}(64, 1)$ \\
             \bottomrule
        \end{tabular} \caption*{}
    \end{subtable}
    \caption{Details on the \textsc{PPO} hyperparameters (\textit{left}) and \textit{actor-critic} architecture (\textit{right}) used in the \textsc{Adverarial-SIPD} experiments. Policy and critic components, and the optimizer, are provided by \textsc{Pytorch}~\citep{paszke2019pytorch}. Implementation and default hyperparameters are derived from \textsc{CleanRL}~\citep{huang2022cleanrl}. These details remained constant throughout the rest of the experiments.} \label{tab:sipd}
\end{table}

\newpage
\section{Extended Experiments Results}
\label{sec:extended-expt-results}

\subsection{Value-Proportional Curriculum}
\begin{figure}[h!]
    \begin{subfigure}{\textwidth}
        \centering
        \begin{subfigure}{0.3\textwidth}
            \centering
            \includegraphics[width=\textwidth]{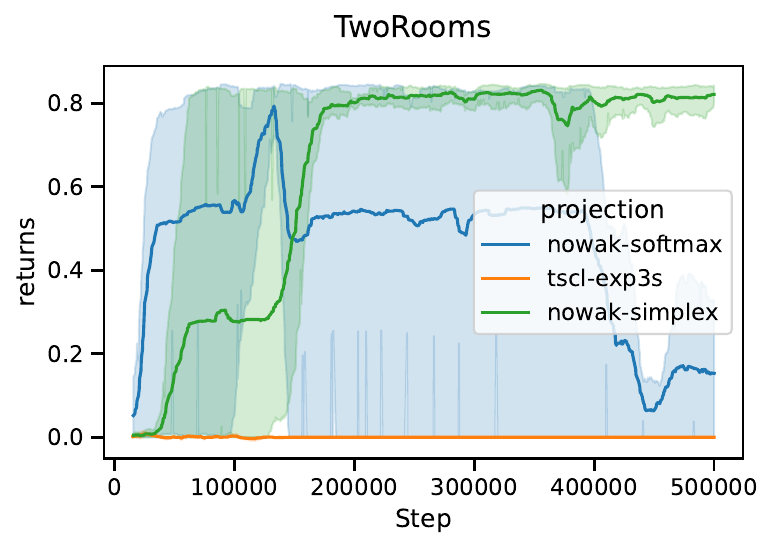}
        \end{subfigure}
        \begin{subfigure}{0.3\textwidth}
            \centering
            \includegraphics[width=\textwidth]{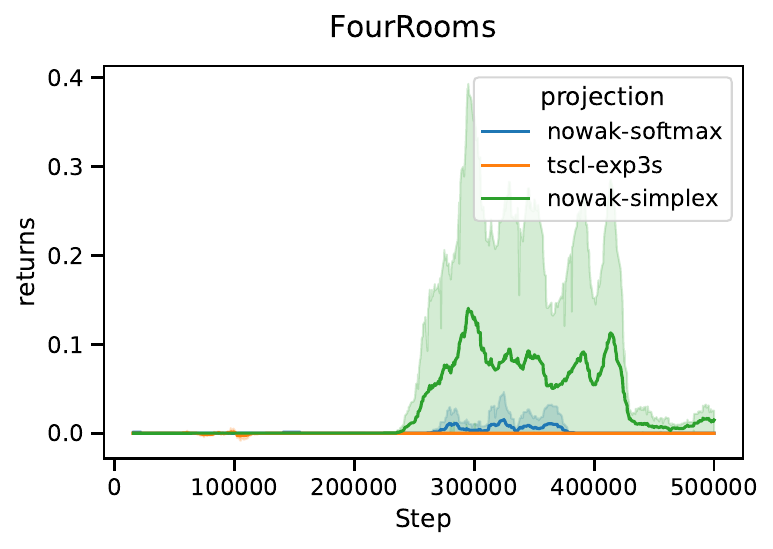}
        \end{subfigure}
        \begin{subfigure}{0.3\textwidth}
            \centering
            \includegraphics[width=\textwidth, height=3.21cm]{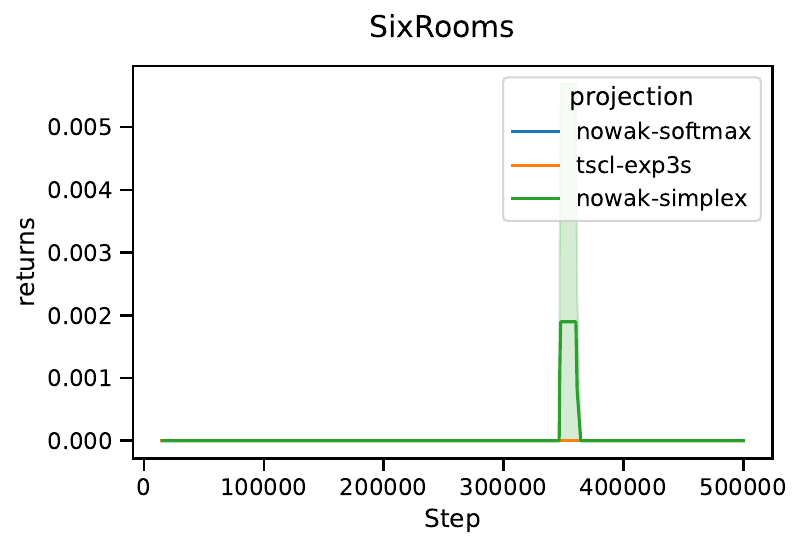}
        \end{subfigure}
        % \begin{subfigure}{0.245\textwidth}
        %     \centering
        %     \includegraphics[width=\textwidth, height=2.25cm]{figures/experiments/propt/Minigrid_Best.pdf}
        % \end{subfigure}
        \caption{\tiny \textsc{MinigridRooms}}
    \end{subfigure}
    \begin{subfigure}{\textwidth}
        \centering
        \begin{subfigure}{0.3\textwidth}
            \centering
            \includegraphics[width=\textwidth]{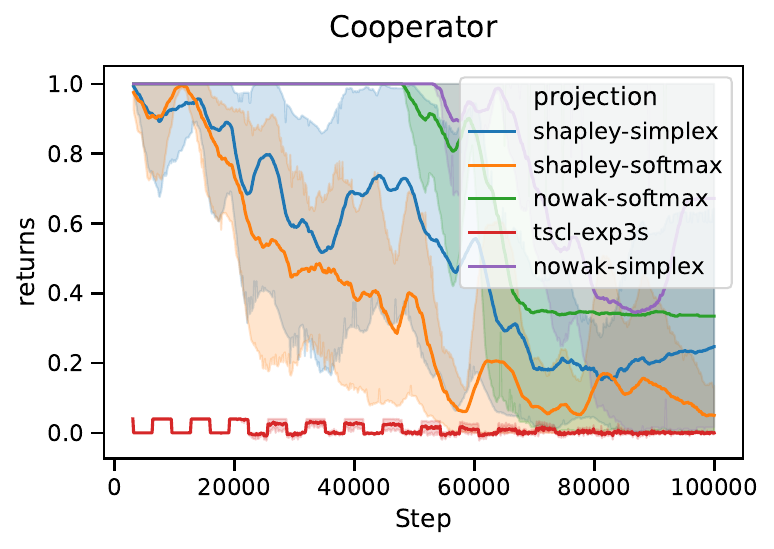}
        \end{subfigure}
        \begin{subfigure}{0.3\textwidth}
            \centering
            \includegraphics[width=\textwidth]{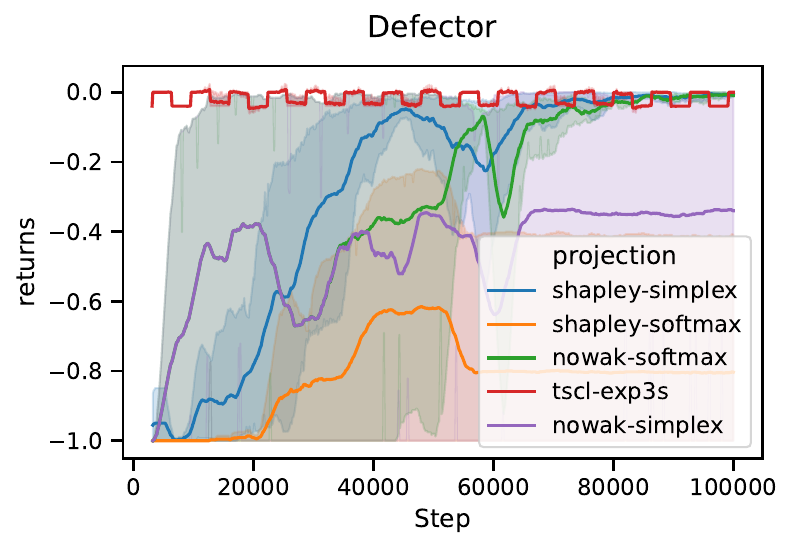}
        \end{subfigure}
        \begin{subfigure}{0.3\textwidth}
            \centering
            \includegraphics[width=\textwidth]{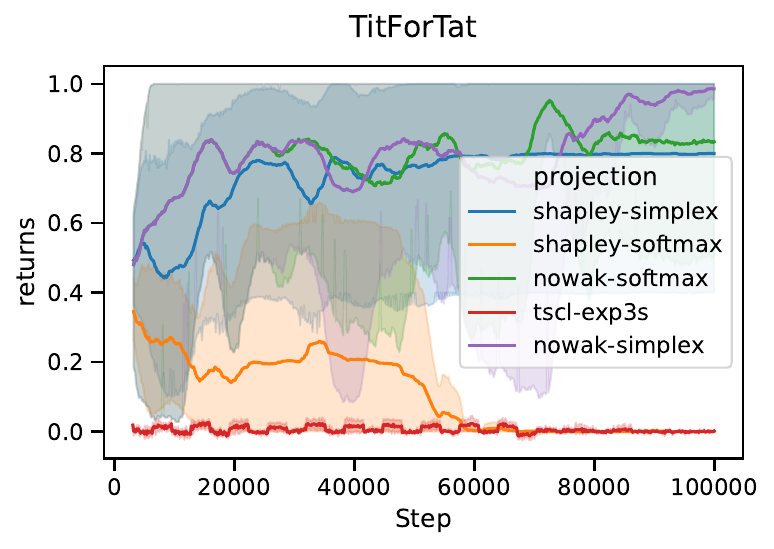}
        \end{subfigure}
        \begin{subfigure}{0.3\textwidth}
            \centering
            \includegraphics[width=\textwidth]{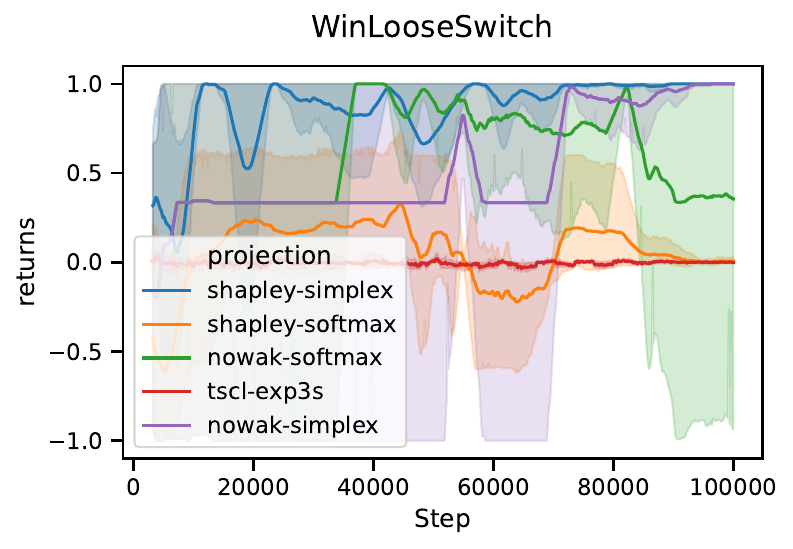}
        \end{subfigure}
        \begin{subfigure}{0.3\textwidth}
            \centering
            \includegraphics[width=\textwidth]{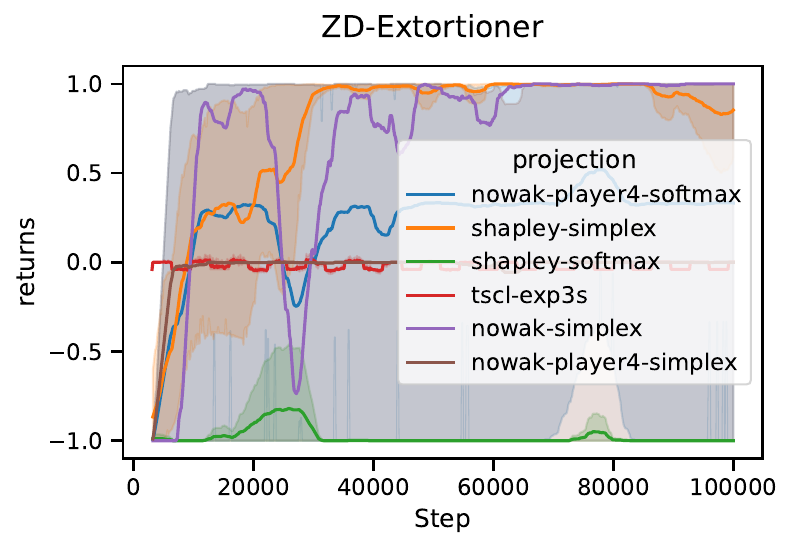}
        \end{subfigure}
        \caption{\tiny \textsc{Adversarial-SIPD}}
    \end{subfigure}
\caption{We also investigated the \emph{prior-proportional curriculum} in the \emph{target-unit} setting. For each target unit, we allocate to each training unit interactions proportional to their pre-computed values for each target. For the \textsc{Adversarial-SIPD} and \textsc{MiniGrid-Rooms} controlled their learning dynamics by presenting the units according to \textit{unordered} and \textit{ordered} mechanisms in~\autoref{sec:expensive-prior}. On each  task, the \emph{value-proportional curriculum} derived from the \emph{prospect priot} outperforms~\ac{TSCL} \emph{(tscl-*-exp3s)}. We further investigate the reason for \ac{TSCL} failures on this scenario.}
\end{figure}

\newpage
\subsection{TSCL Failures}

\begin{figure}[!htb]
    \begin{subfigure}{\textwidth}
        \centering
        \begin{subfigure}{0.3\textwidth}
            \centering
            \includegraphics[width=\textwidth]{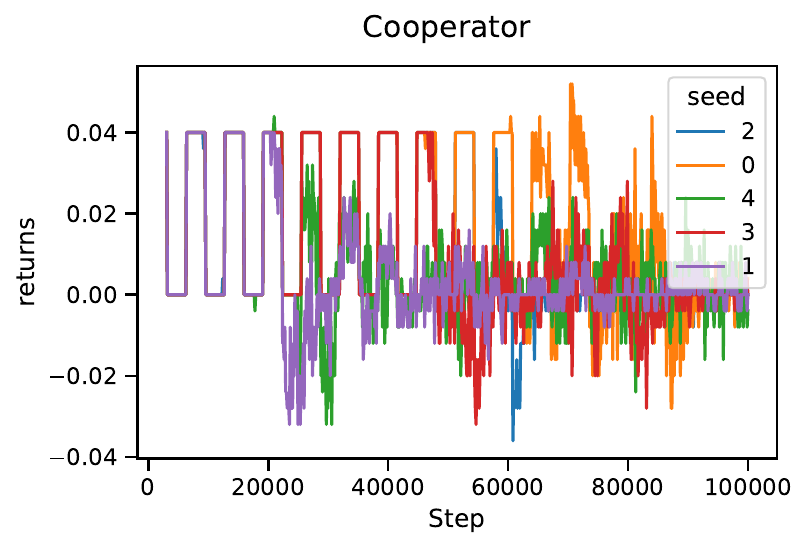}
        \end{subfigure}
        \begin{subfigure}{0.3\textwidth}
            \centering
            \includegraphics[width=\textwidth]{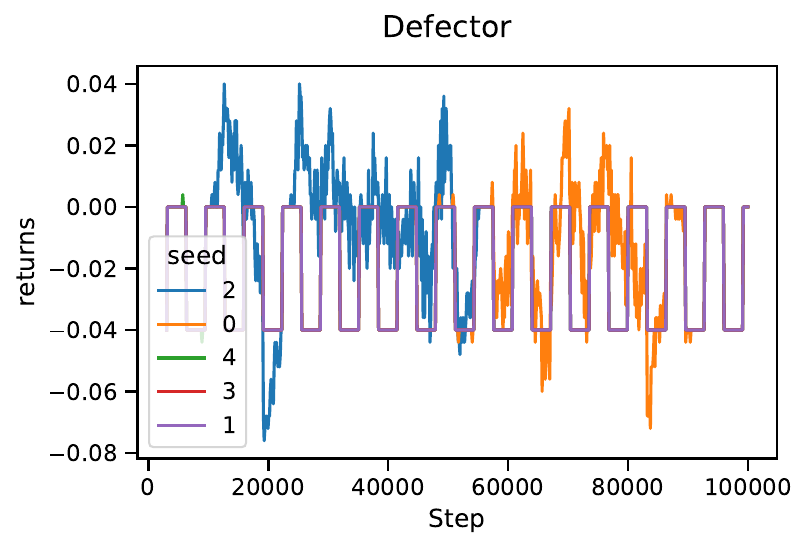}
        \end{subfigure}
        \begin{subfigure}{0.3\textwidth}
            \centering
            \includegraphics[width=\textwidth]{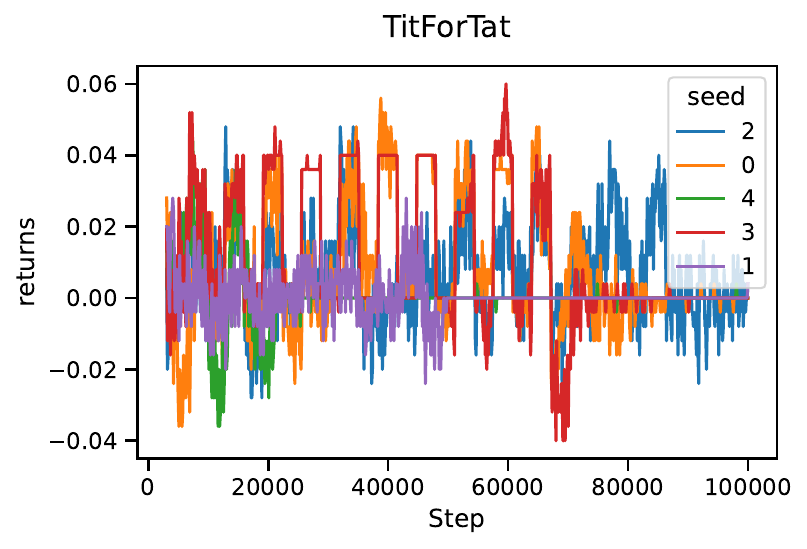}
        \end{subfigure}
        \begin{subfigure}{0.3\textwidth}
            \centering
            \includegraphics[width=\textwidth]{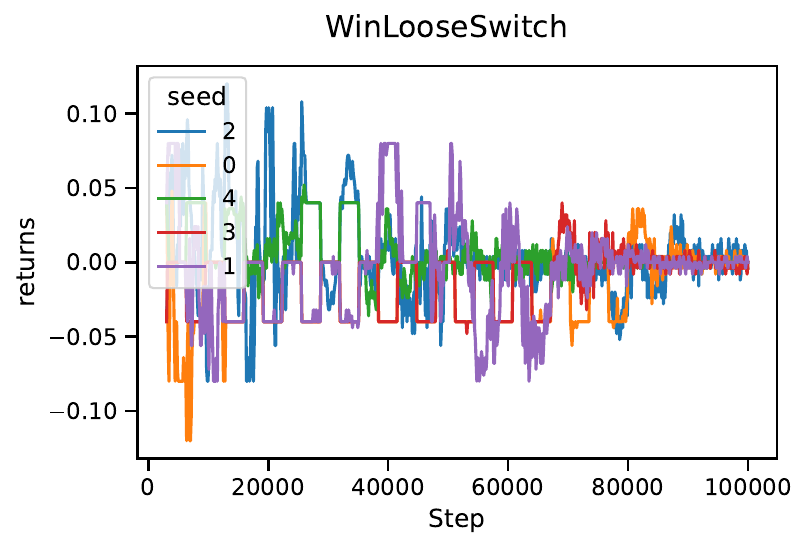}
        \end{subfigure}
        \begin{subfigure}{0.3\textwidth}
            \centering
            \includegraphics[width=\textwidth]{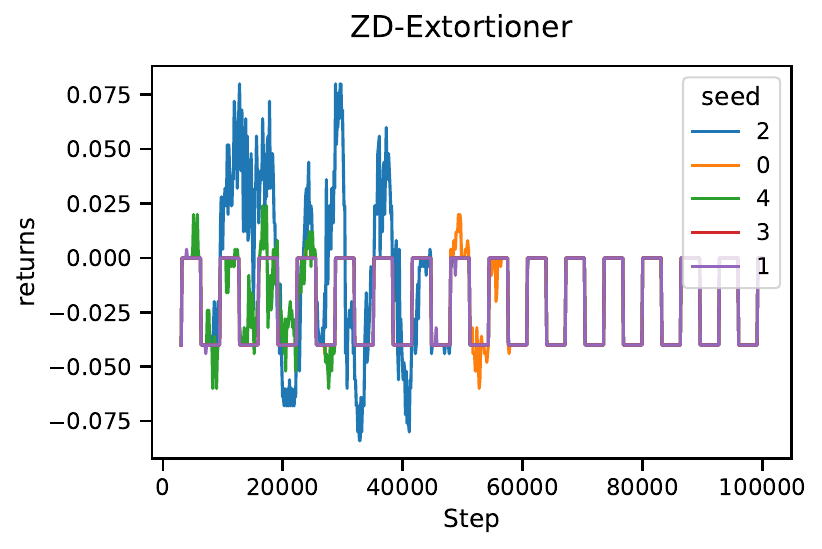}
        \end{subfigure}
        \caption{Individual Runs}
    \end{subfigure}
    \begin{subfigure}{\textwidth}
        \centering
        \begin{subfigure}{0.3\textwidth}
            \centering
            \includegraphics[width=\textwidth]{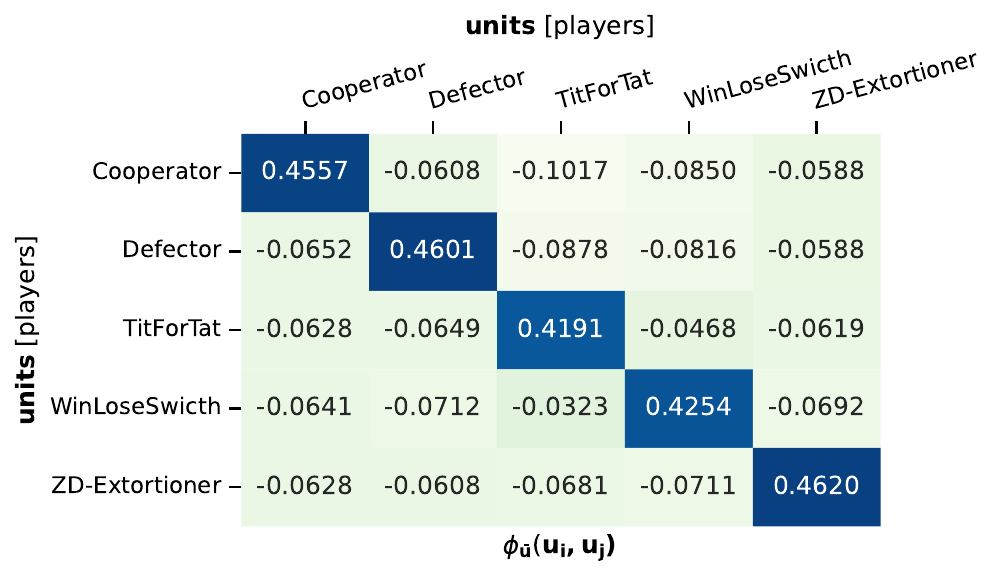}
        \end{subfigure}
        \begin{subfigure}{0.3\textwidth}
            \centering
            \includegraphics[width=\textwidth]{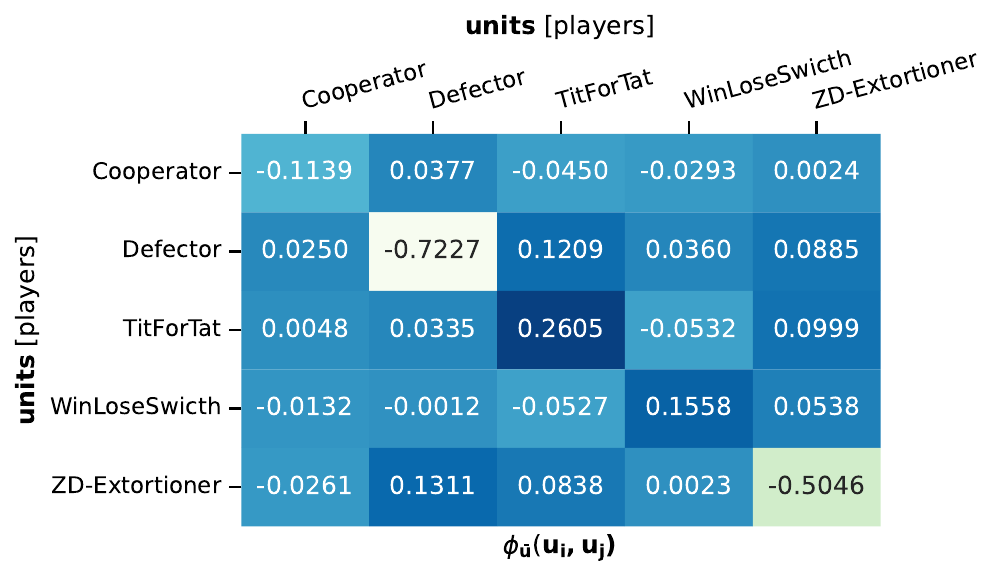}
        \end{subfigure}
        \begin{subfigure}{0.3\textwidth}
            \centering
            \includegraphics[width=\textwidth]{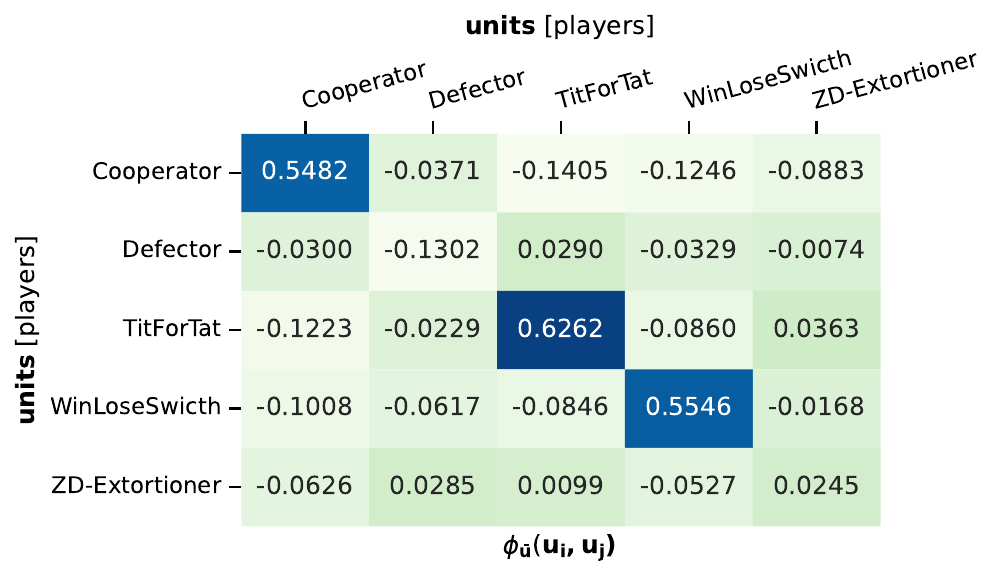}
        \end{subfigure}
        \begin{subfigure}{0.3\textwidth}
            \centering
            \includegraphics[width=\textwidth]{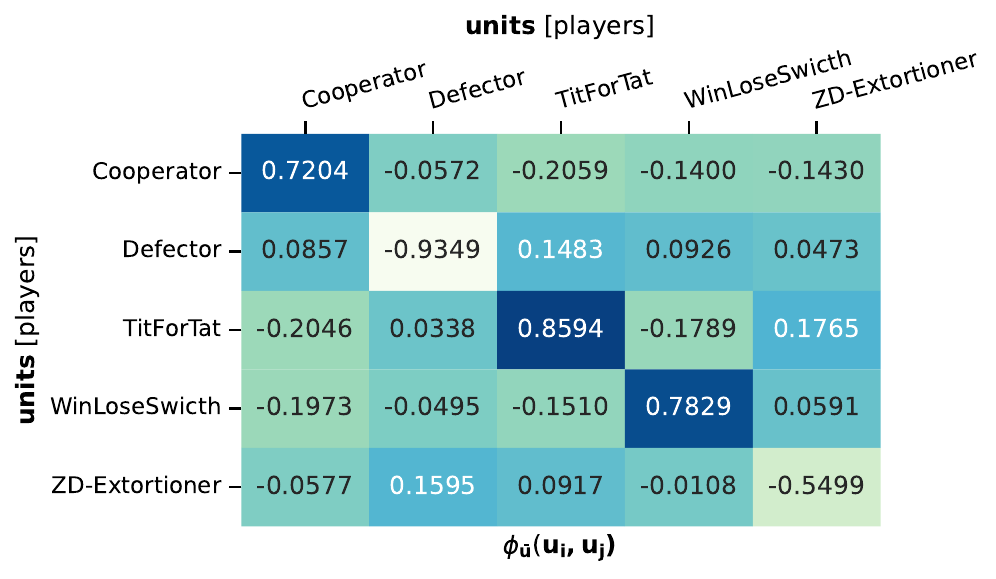}
        \end{subfigure}
        \begin{subfigure}{0.3\textwidth}
            \centering
            \includegraphics[width=\textwidth]{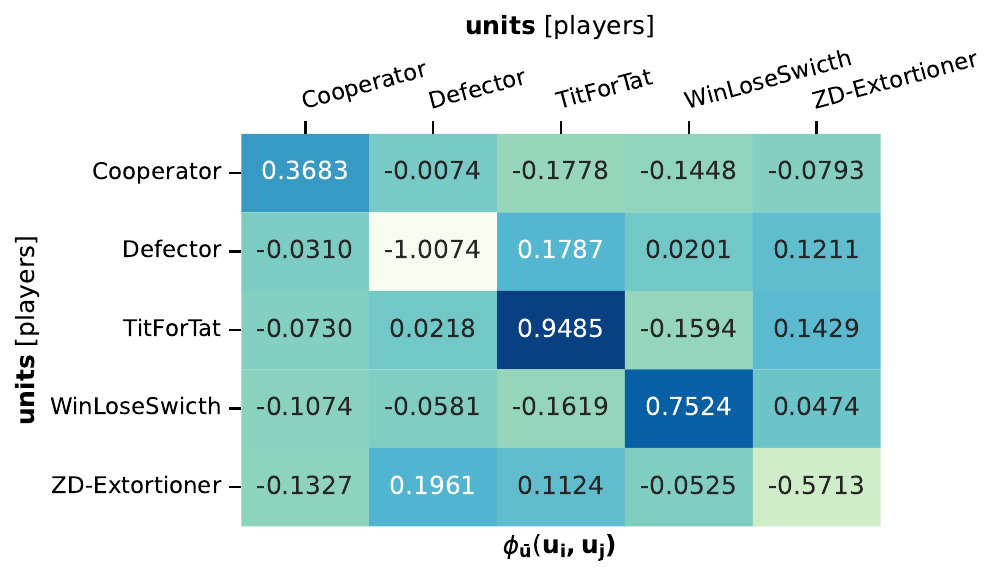}
        \end{subfigure}
        \caption{\tiny \textsc{vPoP Metric}}
    \end{subfigure}
\caption{To understand the failure modes of \ac{TSCL} on \textsc{Adversarial-SIPD}, we represented the individual runs (i.e., each of the five seeds) on every target unit. ~\ac{TSCL} \emph{(tscl-*-exp3s)} (top row) is extremely brittle, unstable, and generally not robust to units interference. We surmise that these failures are related to the \textit{exploration-exploitation} dilemma. Exploratory steps presenting a negatively-valued unit are hard to overcome (forgetting dynamics). This issue requires further investigation, and we defer it to future work. }
\end{figure}

\begin{figure}[t]
    \begin{subfigure}{\textwidth}
        \centering
        \begin{subfigure}{0.3\textwidth}
            \centering
            \includegraphics[width=\textwidth]{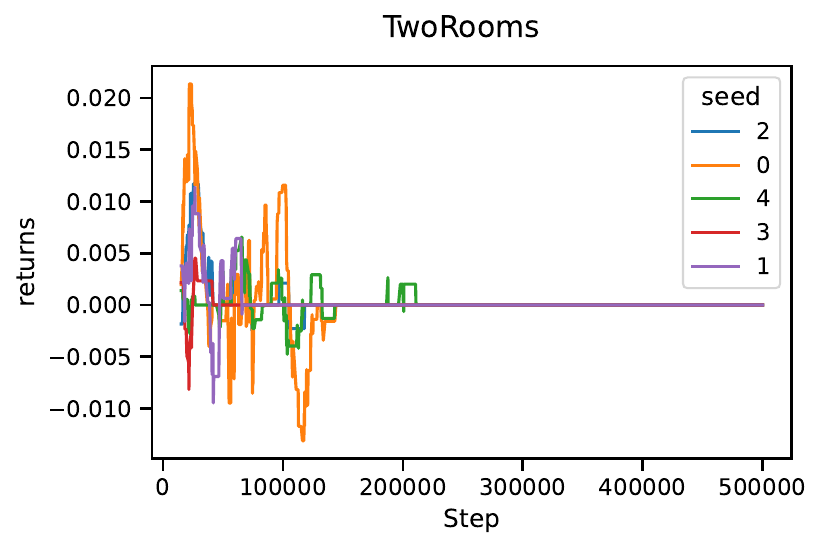}
        \end{subfigure}
        \begin{subfigure}{0.3\textwidth}
            \centering
            \includegraphics[width=\textwidth]{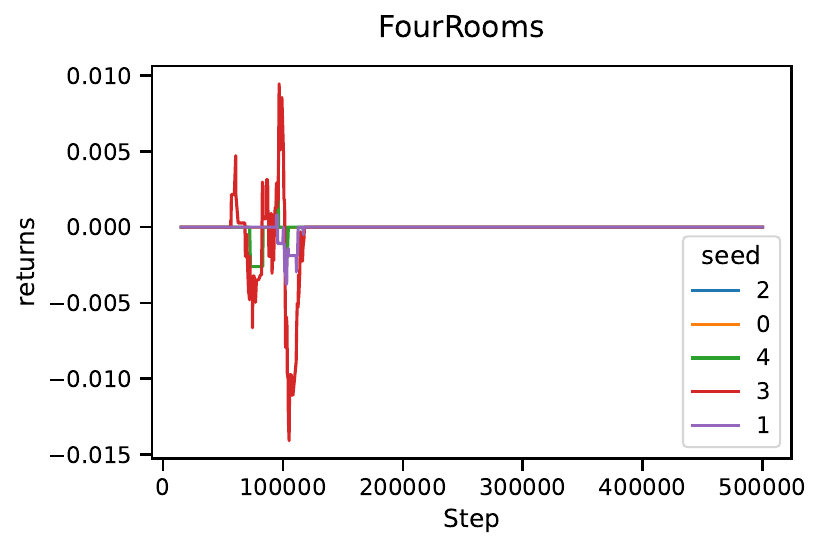}
        \end{subfigure}
        \begin{subfigure}{0.3\textwidth}
            \centering
            \includegraphics[width=\textwidth]{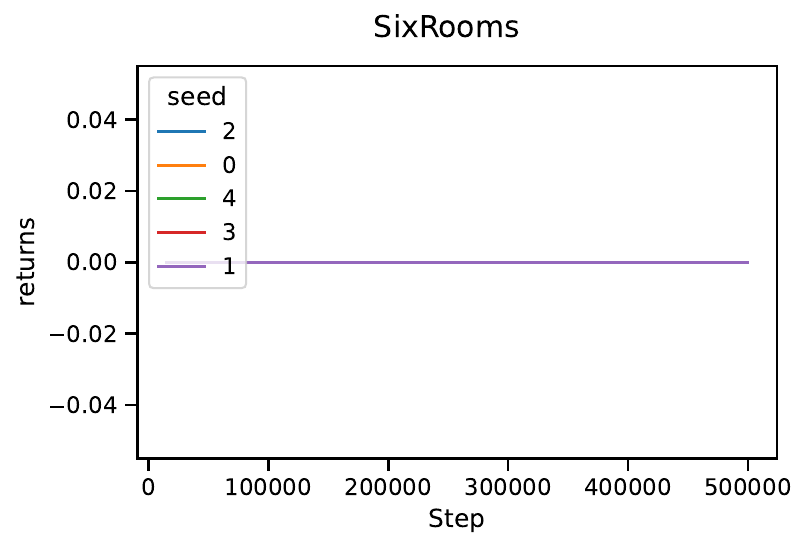}
        \end{subfigure}
        
        \begin{subfigure}{0.3\textwidth}
            \centering
            \includegraphics[width=\textwidth]{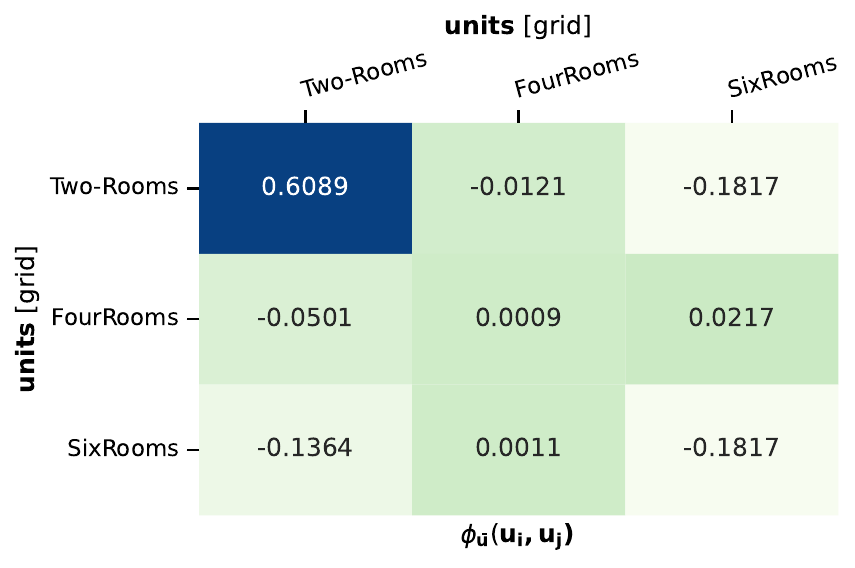}
        \end{subfigure}
        \begin{subfigure}{0.3\textwidth}
            \centering
            \includegraphics[width=\textwidth]{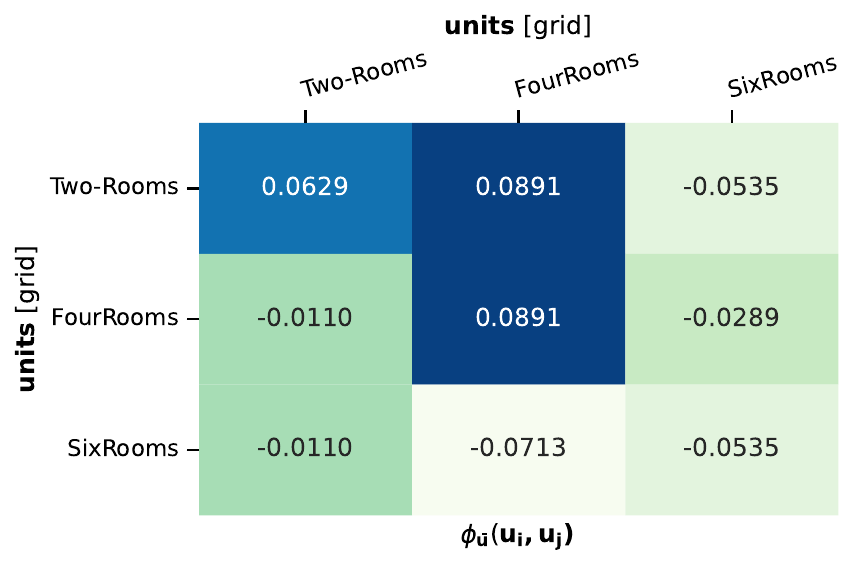}
        \end{subfigure}
        \begin{subfigure}{0.3\textwidth}
            \centering
            \includegraphics[width=\textwidth]{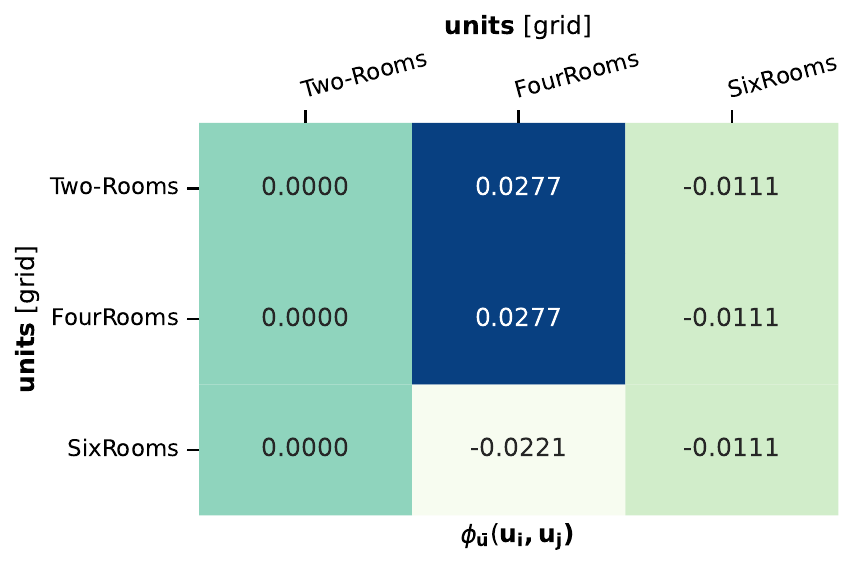}
        \end{subfigure}
        \caption{\tiny \textsc{MinigridRooms}}
    \end{subfigure}
\caption{We found that \ac{TSCL} presents a similar problem in \textsc{MinigridRooms}. When actions (units) need to be almost deterministically drawn for several steps, and other actions (units) have negative interference with the target, \ac{TSCL} is unable to find a stable and robust solution to the p roblem.}
\end{figure}
% \begin{figure}
%     \centering
%     \small
 
%     \begin{subfigure}{0.45\textwidth}
%         \centering
%         \includegraphics[width=\textwidth, height=2.55cm]{figures/experiments/vpop/ASIPD-ONR.pdf}
%         \caption{\tiny \textsc{Adv-SIPD}~\textbf{[vPoP-ordered]}}
%     \end{subfigure}
%     \begin{subfigure}{0.245\textwidth}
%         \centering
%         \includegraphics[width=\textwidth, height=2.55cm]{figures/experiments/vpop/MinAtar.pdf}
%         \caption{\tiny \textsc{MinAtar-1m}~\textbf{[vPoP]}}
%     \end{subfigure}
%     \begin{subfigure}{0.245\textwidth}
%         \tiny
%         \includegraphics[width=\textwidth, height=2.55cm]{figures/experiments/vpop/MinAtar-OSB.pdf}
%         \caption{\tiny \textsc{MinAtar-1m}~\textbf{[vPoP-ordered]}}    
%     \end{subfigure}
%     \caption{\small Experience interference is a universal problem that occurs at any level of aggregation. The Value of a Player to Another Player (vPoP)~\citep{hausken_mohr_2001shapleyothers} captures the pairwise interferences among units of experience. }
%     \label{fig:vpop}
% \end{figure}

% \newpage
% \bibliography{main}
% \bibliographystyle{unsrtnat}

\end{document}